\documentclass{ifacconf}

\usepackage{abbreviations-paper}
\usepackage[utf8]{inputenc}
\usepackage{natbib}
\usepackage{graphics}
\usepackage{epsfig} 
\usepackage[keeplastbox]{flushend}
\usepackage{amsmath} 
\usepackage{amssymb}  
\usepackage{bm}
\usepackage{subcaption}
\usepackage{textcomp}
\usepackage[ruled]{algorithm2e}
\usepackage{url}
\usepackage{xcolor}
\usepackage[labelfont=bf]{caption}
\usepackage{graphicx}
\usepackage{theorem}
\usepackage{booktabs}
\usepackage{dsfont}
\usepackage{booktabs}

\newcommand{\Ts}{\text{s}}
\newcommand{\Tx}{\text{x}}
\newcommand{\Ty}{\text{y}}
\newcommand{\Tr}{\text{r}}

\newcommand{\Tt}{\text{t}}

\newcommand{\qr}{\vx_\text{r}}
\newcommand{\qs}{\vx_\text{s}}
\newcommand{\sbbmat}{\begin{smallmatrix}}
	\newcommand{\sebmat}{\end{smallmatrix}}
\newcommand{\sinc}{\text{sinc}}
\newcommand{\Log}{\text{Log}}
\newcommand{\Exp}{\text{Exp}}
\newcommand{\grad}{\texttt{grad}\,}
\newcommand{\hess}{\texttt{Hess}\,}
\newcommand{\uvx}{\tilde{\vx}}
\newcommand{\ualpha}{\tilde{\bm{\alpha}}}
\newcommand{\valpha}{\bm{\alpha}}
\newcommand{\SET}{\text{SE(}2\text{)}}
\newcommand{\SETT}{\text{SE(}3\text{)}}
\newcommand{\SOT}{\text{SO(}2\text{)}}
\newcommand{\mQ}{\mathcal{Q}}

\newcommand{\mP}{\mathcal{P}}
\newcommand{\mA}{\mathcal{A}}
\newcommand{\mB}{\mathcal{B}}
\newcommand{\mF}{\mathcal{F}}
\newcommand{\mH}{\mathcal{H}}

\SetCommentSty{mycommfont}

\begin{document}
	\begin{frontmatter}
		
		\title{Improved Pose Graph Optimization\\ for Planar Motions\\ Using Riemannian Geometry\\ on the Manifold of Dual Quaternions} 
		
		\thanks[footnoteinfo]{This work is supported by the German Research Foundation (DFG) under grant HA 3789/16-1. Johannes Cox is now with the Technical University of Berlin.}
		
		\author{Kailai~Li,}
		\author{Johannes~Cox,}
		\author{Benjamin~Noack,}
		\author{and Uwe~D.~Hanebeck} 
		
		\address{Intelligent Sensor-Actuator-Systems Laboratory (ISAS) \\Institute for Anthropomatics and Robotics \\Karlsruhe Institute of Technology (KIT), Germany \\kailai.li@kit.edu, j.cox@campus.tu-berlin.de\\ benjamin.noack@kit.edu,  uwe.hanebeck@kit.edu}
		
		\begin{abstract}               
			We present a novel Riemannian approach for planar pose graph optimization problems. By formulating the cost function based on the Riemannian metric on the manifold of dual quaternions representing planar motions, the nonlinear structure of the SE(2) group is inherently considered. To solve the on-manifold least squares problem, a Riemannian Gauss--Newton method using the exponential retraction is applied. The proposed Riemannian pose graph optimizer (RPG-Opt) is further evaluated based on public planar pose graph data sets. Compared with state-of-the-art frameworks, the proposed method gives equivalent accuracy and better convergence robustness under large uncertainties of odometry measurements.
		\end{abstract}
		
	\end{frontmatter}
	
	\section{Introduction}\label{sec:introduction}
	Pose graph optimization (\cite{grisetti2010tutorial}) plays a fundamental role in robotic and computer vision tasks such as simultaneous localization and mapping (SLAM), structure from motion (SfM), etc. A pose graph is a graph with nodes representing robot poses and edges linking the nodes, between which odometry measurements are available. Pose graph optimization problems are typically formulated to maximize the likelihood of the observed odometry measurements of all the edges with constraints imposed by the group structure. As pointed out in~\cite{carlone2014angular}, they are non-convex and contain multiple local minima.  Mathematically speaking, the planar rigid motions belong to the special Euclidean group $\SET$. The frequently used representation approach in the context of planar pose graph optimization is the two-dimensional rotation matrix $\fR\in\SOT$ plus the translation $\vt$\,. Consequently, the three-DoF planar rigid motions are overparameterized by six elements, which leads to three degrees of redundancy. The cost function for performing the maximum likelihood estimation (MLE) is normally formulated as the sum of individual rotational and translational error, which lacks consideration of the underlying group structure. This can be problematic when the rotation and translation errors are at different scales of uncertainty and a proper scaling ratio of the two error terms is not available. Moreover, for highly nonlinear and uncertain rigid motions, the structure-unaware formulations and conventional solvers for constrained optimization problems may have the risk of non-convergence and are prone to local minima. 
	
	The planar dual quaternions are an alternative approach for representing planar rigid motions and can be written as four-dimensional vectors with only one degree of redundancy. Recently, there have been works dedicated to geometry-aware stochastic filters for pose estimation using the (dual) quaternion representation. In~\cite{TAC16_Gilitschenski}, the Bingham distribution was deployed to stochastically model the uncertainty of unit quaternions and corresponding stochastic filters were proposed for orientation estimation. In~\cite{Srivatsan2018} and~\cite{Fusion18_Li-SE2}, dual quaternion-based filtering approaches were proposed for rigid body motion estimation. In~\cite{Cheng2016}, the pose graph optimization problem was re-parameterized by unit dual quaternions and better computational efficiency was shown. However, the deployed objective function heuristically removes entries that might cause singularities without theoretically sound adaptation to the manifold structure of unit dual quaternions. Moreover, the error term in the objective function was merely reformulated from the conventional homogeneous matrix-based scheme. Therefore, the resulting graph optimization scheme was not observed with improved accuracy and robustness, especially in the case of large odometry noise. 
	
	From the perspective of differential geometry, the set of planar dual quaternions forms a Riemannian manifold (\cite{carmo1992riemannian,busam2017camera}) equipped with the so-called \textit{Riemannian metric}, which allows measuring on-manifold uncertainty in accordance with the manifold structure. Furthermore, optimization approaches proposed on Riemannian manifolds, or equivalently \textit{Riemannian optimization} approaches, have been gaining in popularity over the years (\cite{absil2009optimization}). Compared to conventional solvers for constrained optimization problems that use,  e.g., the Lagrange multipliers, the family of Riemannian optimizers have shown better convergence efficiency and robustness as they exploit the geometric structure of the manifold. 
	
	In this paper, we propose the so-called Riemannian pose graph optimizer (RPG-Opt) for planar motions represented by dual quaternions. Here, odometry errors of edges are measured based on the Riemannian metric on the manifold of planar dual quaternions. Unlike conventional solvers for constrained optimization problems as given in~\cite{carlone2014fast,carlone2015lagrangian}, we apply the Riemannian Gauss--Newton approach on the manifold with exponential retraction for updates. Both the cost function formulation and optimization approach are geometry-aware and inherently consider the underlying nonlinear structure of the SE(2) group. For evaluation, we compare the proposed approach with state-of-the-art pose graph optimization frameworks (e.g., GTSAM~\cite{dellaert2012factor}, g2o~\cite{kummerle2011g} and iSAM~\cite{Kaess08tro}) based on public planar pose graph data sets under ordinary odometry noise level. We further synthesize new data sets with additionally larger odometry noise for evaluating the robustness. Comparisons show that our approach gives equally accurate results as the state-of-the-art frameworks and gives superior accuracy under large odometry uncertainties.  
	
	The remainder of the paper is as follows. In Sec.~\ref{sec:dqManifold}, we introduce the dual quaternion parameterization for planar rigid motions and the geometric structure of the planar dual quaternion manifold. Sec.~\ref{sec:RPG} introduces the proposed cost function and the on-manifold optimizer, the RPG-opt. We further evaluate the proposed approach regarding accuracy and convergence robustness in Sec.~\ref{sec:eva}. The work is finally concluded in Sec.~\ref{sec:conclusion}.
	
	\section{Planar Dual Quaternion Representation and Manifold Structure}\label{sec:dqManifold}
	\subsection{Planar Rigid Motion Represented by Dual Quaternions} \label{subsec:dq}
	By convention, unit quaternions representing planar rotations are written as $\vr=\cos(\theta/2)+\vk\sin(\theta/2)$\,,with the unit vector $\vk$ indicating the $z$-axis, around which a rotation of angle $\theta$ is performed (\cite{brookshire2013extrinsic}). Any $\vv\in\R^2$ can be rotated to $\vv^\prime$ according to $\vr$ via $\vv^\prime=\vr\otimes\vv\otimes\vr^*$\,, with $\otimes$ denoting the Hamilton product (\cite{hamilton1844ii}) and $\vr^*=\cos(\theta/2)-\vk\sin(\theta/2)$ the conjugate of $\vr$\,. Moreover, the planar rotation quaternions can be reformulated into the following vector form
	\begin{equation}\label{eq:uq}
	\vr=\bbmat\,\cos(\theta/2),\,\sin(\theta/2)\,\ebmat^\top\in\Sbb^1\subset\R^2\,,
	\end{equation}
	which are located on the unit circle on the $x,y$-plane. Given two planar quaternions $ \vr=[\,r_0,r_1\,]^\top$ and $\vs=[\,s_0,s_1\,]^\top$\,, their Hamilton product can also be represented as ordinary matrix-vector multiplication, namely $\vr\otimes\vs=\mQ_\Tr^\llcorner\,\vs=\mQ_\Ts^\lrcorner\,\vr$\,, with
	\begin{equation} \label{eq:qQ}
	\mQ_\Tr^\llcorner=\bbmat
	r_0 & -r_1 \\
	r_1 & ~~r_0\\
	\ebmat\,,\quad
	\mQ_\Ts^\lrcorner=\bbmat
	~~s_0 & s_1 \\
	-s_1 & ~s_0\\
	\ebmat\,.
	\end{equation}
	It is then trivial to confirm that both the left and right matrix representation belong to the two-dimensional rotation group $\SOT$, i.e., $\mQ\,\mQ^\top=\mQ^\top\mQ=\fI\in\R^{2\times2}$ and its determinant $\det(\mQ)=1$\,.
	The dual quaternion representation for planar rigid motions is defined as $\vx=\vr+\frac{\epsilon}{2}\vt\otimes\vr\,,$ with $\epsilon$ denoting the \textit{dual number} which satisfies $\epsilon^2=0$ and $\vt$ the translation. The corresponding vector form of planar dual quaternions can then be written as 
	\begin{equation*}
	\vx=[\,\qr^\top,\qs^\top\,]^\top\in\Sbb^1\times\R^2:=\mathbb{M}\,,
	\end{equation*}
	with the \textit{real part} $\qr$ defined as in (\ref{eq:uq}) and the \textit{dual part} 
	\begin{equation} \label{eq:dual}
	\qs=\frac{1}{2}\vt\otimes\qr=\frac{1}{2}\mQ^\lrcorner_\Tr\,\vt\,.
	\end{equation}
	Therefore, the manifold of planar dual quaternions (denoted as $\mathbb{M}$) is  the Cartesian product of the unit circle $\Sbb^1$ and the two-dimensional Euclidean space $\R^2$. The arithmetics of dual quaternions are the combination of the Hamilton product and the dual number theory. For instance, given two planar dual quaternions $\vx=[\,x_0,\,x_1,\,x_2,\,x_3\,]^\top$ and $\vy=[\,y_0,\,y_1,\,y_2,\,y_3\,]^\top$\,,  their product in the form of matrix-vector multiplication can be written as $\vx\boxplus\vy=\mQ_\Tx^\ulcorner\,\vy=\mQ_\Ty^\urcorner\,\vx$\,, with
	\begin{equation*} 
	\mQ_\Tx^\ulcorner=\Bigg[\sbbmat
	x_0 &-x_1 & ~~0 & ~~0\\
	x_1 & ~~x_0 & ~~0 & ~~0\\
	x_2 & ~~x_3 & ~~x_0 & -x_1\\
	x_3 & -x_2 &  ~~x_1 &  ~~x_0
	\sebmat\Bigg]\,,\quad
	\mQ_\Ty^\urcorner=\Bigg[\sbbmat
	y_0 & -y_1 & ~~0& ~~0\\
	y_1 & ~~y_0 &  ~~0 & ~~0\\
	y_2 & -y_3 &  ~~y_0 &  ~~y_1\\
	y_3 &  ~~y_2 & -y_1 &  ~~y_0
	\sebmat\Bigg]\,.
	\end{equation*}
	Similar to the rotation rule of unit quaternions, any $\vv\in\R^2$ can be transformed to $\vv^\prime$ via a planar rotation of $\theta$ followed by a translation $\vt$ according to
	\begin{equation*}
	\vv^\prime=\vx\boxplus\vv\boxplus\vx^*\,,
	\end{equation*}
	with $\vx^*=\diag\,(1,-1,-1,-1)\cdot\vx$ being the conjugate of $\vx$\,. Here, $\diag(\cdot)$ is a diagonal matrix with the arguments placed at the diagonal entries. It should be noted that the vector form of dual quaternions for planar motions is expressed w.r.t. the coordinate system $\{1,\vk,\vi,\vj\}$\,, which is a reordering of the general quaternion basis $\{1,\vi,\vj,\vk\}$\,. 
	
	\subsection{Geometry of Planar Dual Quaternion Manifold} \label{subsec:dqStricture} 
	The manifold of planar dual quaternions $\mathbb{M}$ can be derived as the zero-level set of the following vector function defined in $\R^4$
	\begin{equation*}
	g(\vx)=\bbmat\qr^\top\qr-1\\
	\fzero\ebmat,\,\vx=[\,\qr,\qs\,]^\top\in\R^4\,,
	\end{equation*} 
	with $\fzero\in\R^2$ being a zero vector. This can be further used to derive the \textit{tangent plane} $\mathds{T}_\vx\mathbb{M}$ at $\vx\in\mathbb{M}$ by calculating the null space of the Jacobian of $g(\vx)$ evaluated at $\vx$, i.e.,
	\begin{equation*}
	\mathds{T}_\vx\mathbb{M}=\text{ker}\,(\nabla g)=\text{ker}\,\li(\bbmat\,2\qr^\top&\fzero^\top\\\fzero^\top&\fzero^\top\,\ebmat\ri)\,.
	\end{equation*}
	For any $\vy\in\R^4$, its \textit{orthogonal projection} to $\mathds{T}_\vx\mathbb{M}$ can be obtained via $\mP_\vx\vy$, with $\mP_\vx$ being the projection matrix evaluated at $\vx\in\mathbb{M}$\,. As there only exists the unit norm constraint for the real part, the projection matrix can be derived as the following form as given in~\cite{absil2012projection}
	\begin{equation}
	\label{eq:project}
	\mP_\vx=\bbmat\fI-\qr\qr^\top&\fzero\\\fzero&\fI\ebmat\in\R^{4\times4}\,,
	\end{equation}
	with $\fI,\fzero\in\R^{2\times2}$\,. Note that the projection matrix is symmetric and idempotent, namely $\mP_\vx\mP_\vx=\mP_\vx$\,. 
	
	The logarithm map of unit dual quaternions representing the $\SETT$ states can be obtained via the reparameterization into screw motions of $(\theta,d,\vl,\vm)$ according to the Lie algebra as introduced in~\cite{brookshire2013extrinsic} as well as \cite{busam2017camera}. Here, $\theta$ denotes the screw angle (identical to the rotation angle), $d$ the translation along the screw axis $\vl$ (identical to rotation axis) and $\vm$ the screw moment. The logarithm map at the identity dual quaternion $\mathds{1}$ (a quaternion representing zero rotation and zero translation) is given as
	$\text{Log}_\mathds{1}(\vx)=\frac{\theta}{2}\vl+\frac{\epsilon}{2}(\theta\,\vm+d\,\vl)$\,, with $d=\vt^\top\vl$\,. The screw moment $\vm$ is computed w.r.t. the projected origin $\vc$ on the screw axis, namely $\vm=\vc\times\vl$\,. Consequently, the logarithm map of dual quaternions representing planar rigid motions can be derived as a degenerate case of the spatial motions. Therefore, the projected point $\vc$ is the intersection of the screw axis $\vk$ with the $x,y$-plane and can be computed according to~\cite{busam2017camera} as $\vc=\frac{1}{2}(\vt+\vk\times\vt\cot\frac{\theta}{2})$\,.
	
	Considering that the translation $d$ along the screw axis $\vk$ is zero, we have the logarithm map of planar dual quaternion $\vx=[\,x_0,x_1,x_2,x_3\,]^\top$ derived as $\text{Log}_\mathds{1}(\vx)=\frac{1}{2}\,[\,\theta,\,\theta\,\vm^\top\,]^\top$. 
	As a result, it can be further derived that
	\begin{equation*}
	\begin{aligned}
	\theta\,\vm&=\theta\,\vc\times\vk=\frac{\theta}{2}\big(\vt\times\vk+\vt\cot{\frac{\theta}{2}}\big)\\
	&=\frac{1}{\sinc(\theta/2)}\big(\vt\times\vk\sin\frac{\theta}{2}+\vt\cos\frac{\theta}{2}\big)\\
	&=\frac{1}{\sinc(\theta/2)}\bbmat
	~~\cos(\theta/2) & \sin(\theta/2) \\
	-\sin(\theta/2) & \cos(\theta/2)\\
	\ebmat\,\vt\,.
	\end{aligned}
	\end{equation*}
	According to the definition of the dual part in (\ref{eq:dual}), this can be further derived into a more concise form as 
	\begin{equation*}
	0.5\,\theta\,\vm=\frac{\big[\sbbmat
		~~x_0 & x_1 \\
		-x_1 & x_0\\
		\sebmat\big]\vt}{2\,\sinc(\theta/2)}=\frac{\mQ_{\Tr}^\lrcorner\vt}{2\,\sinc(\theta/2)}=\frac{[\,x_2,\,x_3\,]^\top}{\sinc(\theta/2)}\,.
	\end{equation*}
	Therefore, the logarithm of planar dual quaternion can be written as 
	\begin{equation} 
	\label{eq:logmap}
	\vx_\Tt=\text{Log}_\mathds{1}(\vx)=\frac{1}{\gamma}[\,x_1,\, x_2,\, x_3\,]^\top\in\R^3\,,
	\end{equation}
	with $\gamma=\sinc(0.5\,\theta)$ and $\mathds{1}=[\,1,0,0,0\,]^\top$. Moreover, the logarithm map to the tangent plane at $\vx\in\mathbb{M}$ can be computed according to the parallel transport (\cite{busam2017camera}) on the manifold of planar dual quaternions as 
	\begin{equation} \label{eq:logtrans}
	\Log_\vx(\vy)=\vx\boxplus[\,0,\,\Log_\mathds{1}(\vx^{-1}\boxplus\vy)^\top\,]^\top\in\mathds{T}_\vx\mathbb{M}\,.
	\end{equation}
	Conversely, the exponential map at identity $\mathds{1}$ maps a point $\vx_\Tt=[\,x_{\Tt,1},x_{\Tt,2},x_{\Tt,3}\,]^\top$ from the tangent plane at $\mathds{1}$ back to the manifold via
	\begin{equation*}
	\vx=\Exp_\mathds{1}(\vx_\Tt)=[\,\cos(x_{\Tt,1}),\,\gamma\vx_\Tt^\top\,]^\top\,.
	\end{equation*}
	Here, $\gamma=\sinc(0.5\,\theta)=\sinc(x_{\Tt,1})$ as given in (\ref{eq:logmap}). Similar to the logarithm map, the exponential map for arbitrary tangent plane locations can be derived according to the parallel transport 
	\begin{equation}
	\label{eq:exp}
	\Exp_\vx(\vy_\Tt)=\vx\boxplus\Exp_\mathds{1}\big((\vx^{-1}\boxplus\vy_\Tt)_{1:3}\big)\,.
	\end{equation} 
	Here, $\vy_\Tt\in\mathds{T}_\vx\mathbb{M}$ and we take out the last three indices of $\vx^{-1}\boxplus\vy_\Tt$ as its first element is zero according to (\ref{eq:logtrans}).
	
	\section{RPG-Opt:\protect\\ A Riemannian Pose Graph Optimizer} \label{sec:RPG}
	\subsection{Geometry-Aware Cost Function}
	The pose graph optimization problem is formulated as the maximum likelihood estimation given the observed odometry. The optimized poses on the graph $\mathds{C}$ can then be obtained by minimizing the sum of the distance metrics through each edge of the graph, namely  
	\begin{equation} \label{eq:MAP}
	\tilde{\vx}^* = \argmin_{\tilde{\vx}\in\mathbb{M}^n}{\sum}_{\langle i,j\rangle\in\mathds{C}}\,\ve_{ij}^\top\,\bm{\Omega}_{ij}\,\ve_{ij}\,,
	\end{equation}
	with $\tilde{\vx}\in\R^{4n}$ being the poses of $n$ graph nodes concatenated into one vector, where each pose $\vx_i\in\mathbb{M}\subset\R^4$ is a planar dual quaternion. Here, $\bm{\Omega}_{ij}\in\R^{3\times3}$ denotes the uncertainty of the odometry measurement in the form of information matrix. The manifold for optimization is thus the Cartesian product of the planar dual quaternion manifold $\mathbb{M}$, which is also a Riemannian  manifold according to~\cite{birdal2018bayesian}. The cost function 
	\begin{equation}
	\label{eq:obj}
	\mathcal{F}(\uvx)={\sum}_{\langle i,j\rangle\in\mathds{C}}\,\ve_{ij}^\top\,\bm{\Omega}_{ij}\,\ve_{ij}
	\end{equation}
	is thus a scalar function proposed on the manifold $\mathbb{M}^n$, namely $\mathcal{F}:\mathbb{M}^n\rightarrow\R$\,. Unlike the existing works in~\cite{Kaess09ras},~\cite{Dellaert06ijrr} and~\cite{kummerle2011g}, we propose a cost function based on the Riemannian metric (\cite{absil2009optimization,arvanitidis2016nips}) on the manifold of planar dual quaternions, which inherently considers the geometric structure of the nonlinear manifold. Given the odometry measurement $\vz_{ij}$ of the edge connecting  node $\vx_i$ and $\vx_j$, the error term of the edge is defined as
	\begin{equation}
	\label{eq:error}
	\ve_{ij}=\Log_\mathds{1}(\vz_{ij}^{-1}\boxplus\vx_i^{-1}\boxplus\vx_j)\in\R^3\,,
	\end{equation}
	such that the cost function for a single edge can be derived according to the Riemannian metric as
	\begin{equation*}
	f_{ij}(\vx_i,\vx_j)=\frac{1}{2}\ve_{ij}^\top\,\bm{\Omega}_{ij}\,\ve_{ij}=\frac{1}{2}\Vert\Log_\mathds{1}(\vz_{ij}^{-1}\boxplus\vx_i^{-1}\boxplus\vx_j)\Vert^2_{\bm{\Omega}_{ij}}\,,
	\end{equation*}
	which is the Mahalanobis distance measured on the tangent plane at the identity planar dual quaternion $\mathds{1}$\,. 
	\begin{table}
		\centering
		\renewcommand{\arraystretch}{1.2}
		\newcommand{\eye}{$\bm{I}$}
		\captionsetup{width=0.49\textwidth}
		\caption{Evaluation results under ordinary noise.}
		\begin{tabular}{|p{0.95cm}|p{0.1cm}|p{1.28cm}|p{1.28cm}|p{1.28cm}|p{1.28cm}|p{1.28cm}|}
			\hline
			Dataset &		& RPG-Opt10 		& g2o10			   & GTSAM & iSAM \\ \hline 
			\textit{CSAIL} & \fI & $1.07\cdot10^{-1}$&$1.07\cdot10^{-1}$ & $1.07\cdot10^{-1}$ & $1.07 \cdot 10^{-1}$ \\\cline{2-6}
			& $\bm{\Omega}$	& $3.90 \cdot 10^1$	& $3.90 \cdot 10^1$    & $3.90 \cdot 10^1$ & $9.40 \cdot 10^2$\\ \hline
			
			\textit{FR079} & \fI   & $7.19 \cdot 10^{-2}$ & $7.19 \cdot 10^{-2}$ & $7.19 \cdot 10^{-2}$ & $7.19 \cdot 10^{-2}$ \\\cline{2-6}
			& $\bm{\Omega}$	& $3.76 \cdot 10^1$	& $3.76 \cdot 10^1$    & $3.76 \cdot 10^1$ & $3.50 \cdot 10^2$ \\ \hline
			
			\textit{FRH}	& \fI	   & $1.39 \cdot 10^{-6}$ & $3.19 \cdot 10^{-4}$ & $3.19 \cdot 10^{-4}$ & $3.46 \cdot 10^{-4}$ \\\cline{2-6}
			& $\bm{\Omega}$	& $1.93 \cdot 10^{-4}$ & $4.18 \cdot 10^{-2}$ & $4.18 \cdot 10^{-2}$ & $4.61 \cdot 10^{-2}$ \\ \hline
			
			\textit{M3500} & \fI    & $3.02 \cdot 10^{0}$   & $3.02 \cdot 10^{0}$ & $3.02 \cdot 10^{0}$ & $3.02 \cdot 10^{0}$ \\\cline{2-6}
			& $\bm{\Omega}$ & $1.38 \cdot 10^{2}$ & $1.38 \cdot 10^{2}$  & $1.38 \cdot 10^{2}$ & $1.39 \cdot 10^{2}$\\ \hline
			
			\textit{MITb} & \fI     & $6.60 \cdot 10^{0}$ & $2.70 \cdot 10^{1}$ &	$2.83 \cdot 10^{6}$ &	$2.78 \cdot 10^{0}$\\\cline{2-6}
			& $\bm{\Omega}$ & $3.02 \cdot 10^{3}$ & $7.71 \cdot 10^{2}$  & $4.49 \cdot 10^{9}$ & $2.26 \times 10^{2}$ \\ \hline
			
			\textit{City10K} & \fI &	$8.72\cdot 10^{0}$ & $8.72 \cdot 10^{0}$ & $6.77 \cdot 10^{6}$ & $8.72 \cdot 10^{0}$\\\cline{2-6}
			& $\bm{\Omega}$	& $5.12 \cdot 10^{2}$ & $5.12 \cdot 10^{2}$  & $2.48 \cdot 10^{8}$ & $5.18 \cdot 10^{2}$ \\ \hline
			
			\textit{M10K} & \fI & $3.03\cdot 10^{2}$ & $3.03 \cdot 10^{2}$ & $3.17 \cdot 10^{7}$ & $3.03 \cdot 10^{2}$ \\\cline{2-6}
			& $\bm{\Omega}$	& $1.99 \cdot 10^{5}$ & $1.98 \cdot 10^{5}$  & $2.91 \cdot 10^{12}$ & $2.93 \cdot 10^{8}$ \\
			\hline
		\end{tabular}
		\label{tab:easy}
	\end{table}
	\begin{table*}
		\captionsetup{width=.9\textwidth}
		\centering
		\renewcommand{\arraystretch}{1.5}
		\newcommand{\erot}[2]{$#1 \cdot 10^{#2}$}
		\newcommand{\etrans}[2]{$#1 \cdot 10^{#2}$}
		\caption{Summarized evaluation results for synthetic data sets with chordal relaxation-based initialization.}
		\begin{tabular}{|l|cc|cc|cc|cc|}
			\hline
			& \multicolumn{2}{c|}{RPG-Opt} & \multicolumn{2}{c|}{chord + RPG-Opt} & \multicolumn{2}{c|}{chord + g2o} & \multicolumn{2}{c|}{chord + GTSAM}  \\ \hline
			Dataset& $e_\text{t}$ & $e_\text{r}$ & $e_\text{t}$ & $e_\text{r}$ & $e_\text{t}$ & $e_\text{r}$ & $e_\text{t}$ & $e_\text{r}$ \\  \hline
			\textit{M3500a+} & $2.60 \cdot 10^{-1}$ & $8.42 \cdot 10^{0}$   & $2.60 \cdot 10^{-1}$ & $8.42 \cdot 10^{0}$ & $2.60 \cdot 10^{-1}$ & $8.42 \cdot 10^{0}$ & $3.310 \cdot 10^{0}$ & $1.55 \cdot 10^{1}$ \\
			\textit{M3500b+} & $3.79 \cdot 10^{-1}$ & $1.31 \cdot 10^{1}$  & $3.79 \cdot 10^{-1}$ & $1.31 \cdot 10^{1}$ & $3.80 \cdot 10^{-1}$ & $1.31 \cdot 10^{1}$ & $4.37 \cdot 10^{0}$ & $2.03 \cdot 10^{1}$ \\
			\textit{M3500c+} & \etrans{3.94}{-1} & \erot{1.31}{1} & \etrans{3.94}{-1} & \erot{1.31}{1} & \etrans{3.91}{-1} & \erot{1.31}{1} & \etrans{1.14}{0} & \erot{1.36}{1} \\
			\textit{M3500d+} & \etrans{8.28}{-1} & \erot{1.51}{1} & \etrans{4.71}{-1} & \erot{1.51}{1} & \etrans{4.78}{-1} & \erot{1.51}{1} & \etrans{7.01}{-1} & \erot{1.57}{1} \\
			\textit{City10000a} & $4.34 \cdot 10^{-2}$ & $1.55 \cdot 10^{0}$  & $4.34 \cdot 10^{-2}$ & $1.55 \cdot 10^{0}$ & $4.34 \cdot 10^{-2}$ & $1.55 \cdot 10^{0}$ & $4.33 \cdot 10^{-2}$ & $1.55 \cdot 10^{0}$ \\
			\textit{City10000b} & $3.20 \cdot 10^{-1}$ & $6.95 \cdot 10^{0}$  & $5.32 \cdot 10^{-2}$ & $2.66 \cdot 10^{0}$ & $5.32 \cdot 10^{-2}$ & $2.66 \cdot 10^{0}$ & $5.29 \cdot 10^{-2}$ & $2.66 \cdot 10^{0}$ \\ 
			\textit{City10000c} & \etrans{2.86}{-2} & \erot{5.80}{0} & \etrans{5.77}{-2} & \erot{2.67}{0} & \etrans{5.77}{-2} & \erot{2.67}{0} & \etrans{5.77}{-2} & \erot{2.67}{0} \\
			\textit{City10000d} & \etrans{3.88}{-1} & \etrans{1.39}{1} & \etrans{2.46}{-1} & \erot{1.22}{1} & \etrans{2.46}{-1} & \etrans{1.23}{-1} & \etrans{9.28}{-1} & \erot{1.39}{1} \\ \hline
		\end{tabular}
		\label{tab:sum}
	\end{table*}
	\subsection{Riemannian Gauss--Newton Method for On-Manifold Planar Pose Graph Optimization}
	We apply the Riemannian Gauss--Newton approach in~\cite{absil2009optimization} for solving the nonlinear least square problem formulated in (\ref{eq:MAP}). Here, a new iteration $\uvx^{k+1}$ is obtained by retracting  the on-tangent-plane Newton step back to the manifold, namely
	\begin{equation*}
	\uvx^{k+1} = \mathcal{R}_{\uvx^k}(\ualpha^k)\,,\bm{\ualpha}^k\in\mathds{T}_{\uvx^k}\mathbb{M}^n\,.
	\end{equation*} 
	The iterative step $\bm{\ualpha}^k$ results from the \textit{Riemanian gradient} $\grad F(\uvx^k)$ and the \textit{Riemannian Hessian} $\hess F(\uvx^k)$ via
	\begin{equation*}
	\hess F(\uvx^k)\,\bm{\ualpha}^k=-\grad F(\uvx^k)\,,
	\end{equation*}
	with $\bm{\ualpha}^k$ being the concatenated on-tangent-plane steps calculated for each node. Without loss of generality, we perform the following derivations based on the cost function $f_{ij}$ of one single edge joining $\vx_i$ and $\vx_j$. For better readability, we ignore the iteration index $k$\,.  The Riemannian gradient w.r.t. $\vx_i$ can be computed by orthogonally projecting the ordinary gradient of $f_{ij}$ onto the tangent plane at $\vx_i$ via
	\begin{equation*}
	\grad f_{ij}(\vx_i)=\mP_{\vx_i}\nabla f_{ij}(\vx_i)\,,
	\end{equation*}
	with $\mP_{\vx_i}$ being the projection matrix to $\mathds{T}_{\vx_i}\mathbb{M}$ given in (\ref{eq:project}) and $\nabla f_{ij}(\vx_i)=\mA_{ij}^\top\,\bm{\Omega}_{ij}\,\ve_{ij}$ the classical gradient in the Euclidean space. Here, $\mA_{ij}$ is the classical Jacobian of the error metric $\ve_{ij}$ w.r.t. $\vx_i$\,. Therefore, we have
	\begin{equation*}
	\grad f_{ij}(\vx_i)=\mP_{\vx_i}\mA_{ij}^\top\,\bm{\Omega}_{ij}\,\ve_{ij}\,.
	\end{equation*}
	Similarly, the Riemannian gradient at node $\vx_j$ can be calculated as 
	\begin{equation*}
	\grad f_{ij}(\vx_j)=\mP_{\vx_j}\mB_{ij}^\top\,\bm{\Omega}_{ij}\,\ve_{ij}\,,
	\end{equation*}
	with $\mB_{ij}$ denoting the classical Jacobian of $\ve_{ij}$ w.r.t. $\vx_j$\,. The Riemannian gradient of $\mF$ in \eqref{eq:obj} is then computed by traversing all the nodes through the graph with each entry computed as
	\begin{equation*}
	\begin{aligned}
	\grad \mF_{[i]}&=\mP_{\vx_i}\mA_{ij}^\top\,\bm{\Omega}_{ij}\,\ve_{ij}\\
	\grad F_{[j]}&=\mP_{\vx_j}\mB_{ij}^\top\,\bm{\Omega}_{ij}\,\ve_{ij}\,.
	\end{aligned}
	\end{equation*}
	The Riemannian Hessian can be computed by approximation (\cite{absil2009optimization}) with entries as follows 
	\begin{equation*}
	\begin{aligned}
	&\mH_{[ii]}=\mP_{\vx_i}\mA_{ij}^\top\,\bm{\Omega}_{ij}\,\mA_{ij}\,\mP_{\vx_i}^\top\,,\\
	&\mH_{[ij]}=\mP_{\vx_i}\mA_{ij}^\top\,\bm{\Omega}_{ij}\,\mB_{ij}\,\mP_{\vx_j}^\top\,,\\
	&\mH_{[ji]}=\mP_{\vx_j}\mB_{ij}^\top\,\bm{\Omega}_{ij}\,\mA_{ij}\,\mP_{\vx_i}^\top\,,\\
	&\mH_{[jj]}=\mP_{\vx_j}\mB_{ij}^\top\,\bm{\Omega}_{ij}\,\mB_{ij}\,\mP_{\vx_j}^\top\,,
	\end{aligned}
	\end{equation*}
	Here, $i$ and $j$ indicate the locations of block matrices in the Hessian matrix $\mH$ corresponding to the nodes $\vx_i$ and $\vx_j$\,. 
	Finally, the iterative step can be obtained by solving the linear system
	\begin{equation*}
	\mH\,\ualpha=-\grad\mF\,,
	\end{equation*}
	with $\ualpha$ denoting the on-tangent-plane iteration concatenated through each node.
	
	\begin{figure*}[t]
		\centering
		\newcommand{\erot}[2]{$e_\text{r} = #1 \cdot 10^{#2}$}
		\newcommand{\etrans}[2]{$e_\text{t} = #1 \cdot 10^{#2}$}
		\setlength\tabcolsep{2pt}
		$\bm{\Sigma}$ 
		\begin{tabular}{cccccc}
			\textit{M3500a+}  & \textit{M3500b+} & \textit{M3500c+}  & \textit{City10000a} &  \textit{City10000b} & \textit{City10000c} \\
			$\Big[\sbbmat0.0224&0&0\\0&0.0224&0\\0&0&0.0224\sebmat\Big]$& $\Big[\sbbmat0.0224&0&0\\0&0.0224&0\\0&0&0.1\sebmat\Big]$& $\Big[\sbbmat0.0224&0.01&0.025\\0.01&0.0224&0.025\\0.025&0.025&0.1\sebmat\Big]$& $\Big[\sbbmat0.001&0&0\\0&0.001&0\\0&0&0.002\sebmat\Big]$& $\Big[\sbbmat0.0015&0&0\\0&0.0015&0\\0&0&0.007\sebmat\Big]$& $\Big[\sbbmat0.0015& 0.0005&0.001\\0.0005&0.0015&0.001\\0.001&0.001&0.007\sebmat\Big]$\\
		\end{tabular}\\
		
		\rotatebox[origin=c]{90}{odometry} 
		\begin{tabular}{cccccc}
			\includegraphics[height=0.165\textwidth]{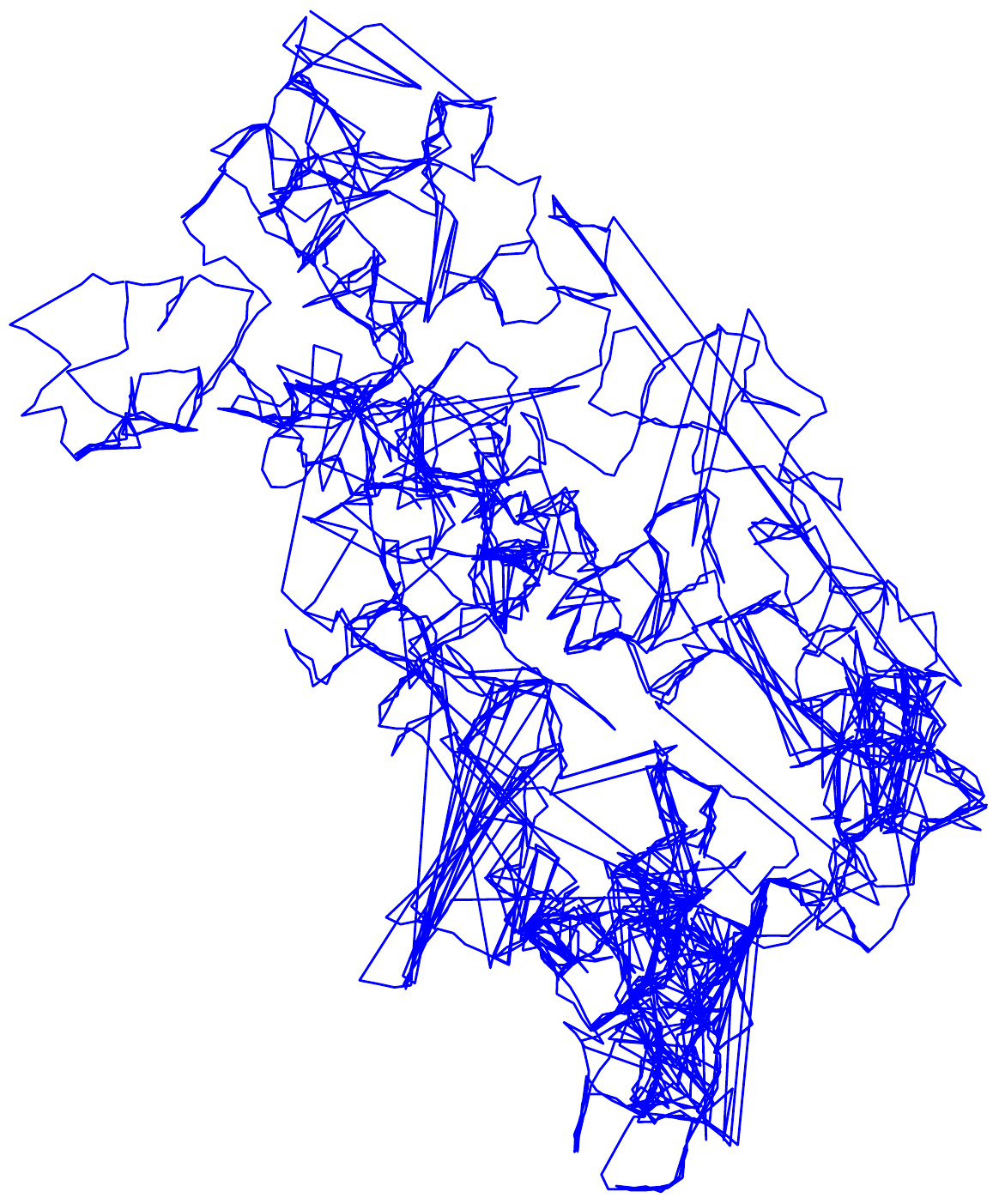}& \includegraphics[width=0.165\textwidth]{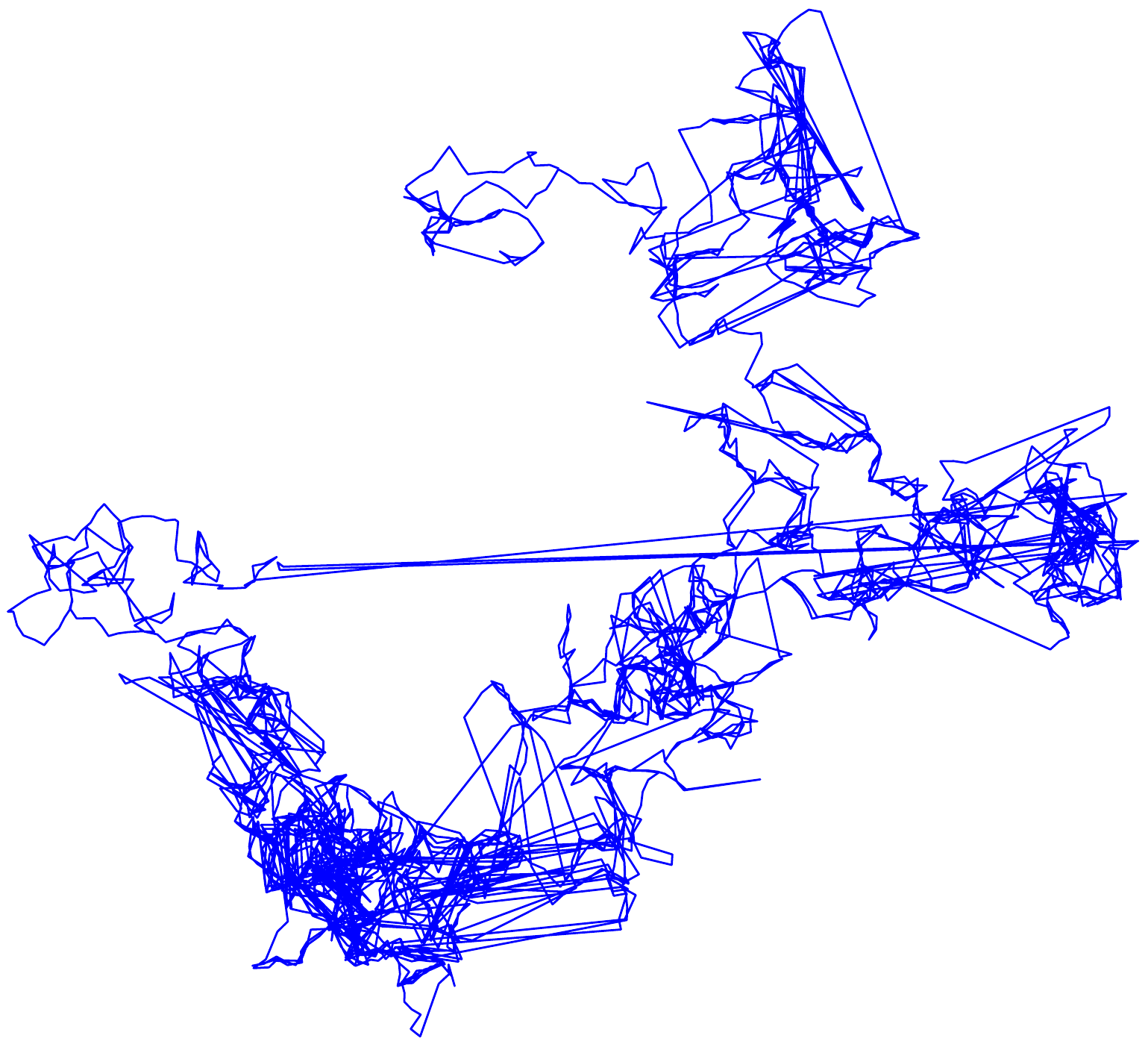}&  \includegraphics[height=0.165\textwidth]{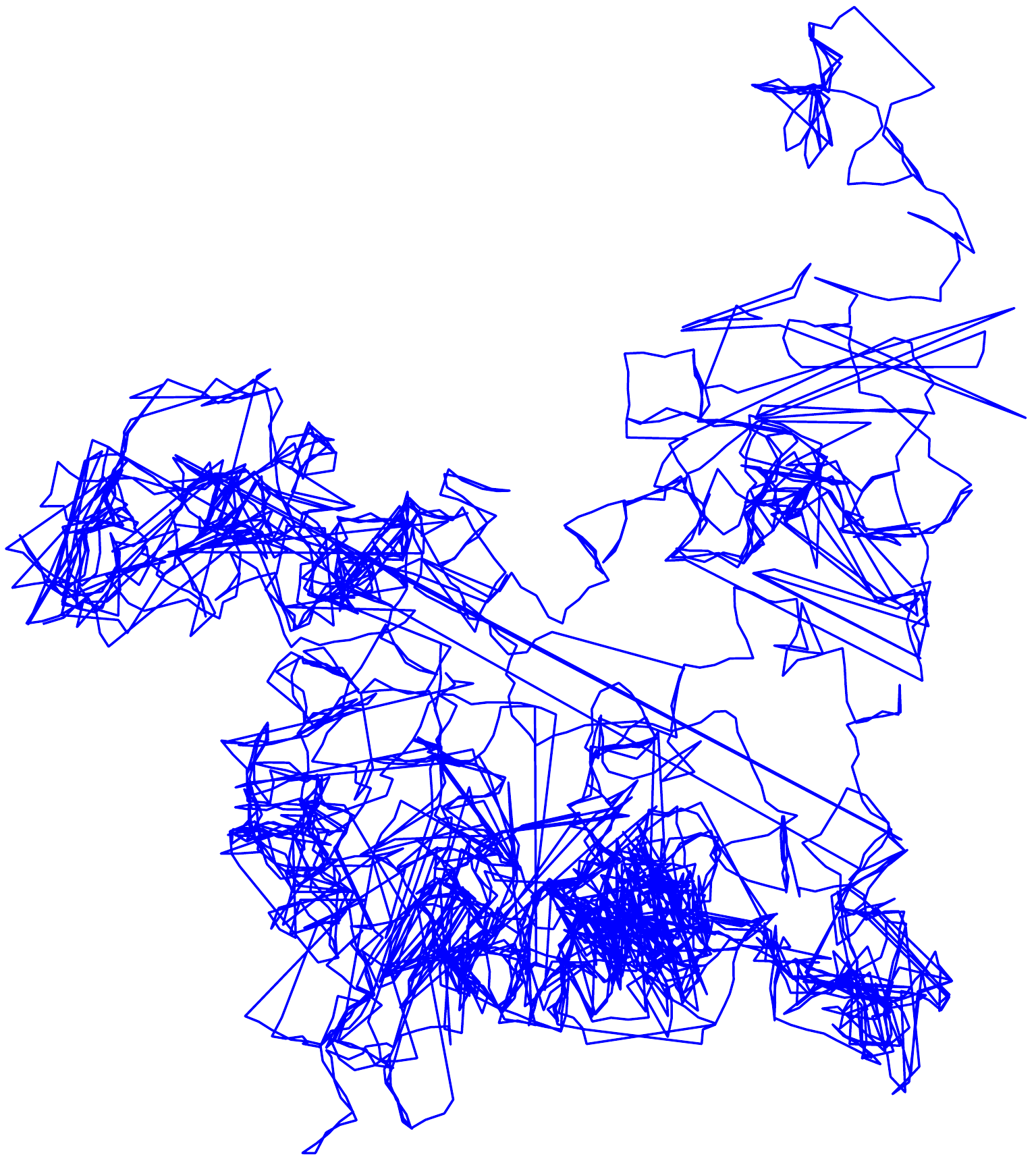}& \includegraphics[width=0.155\textwidth]{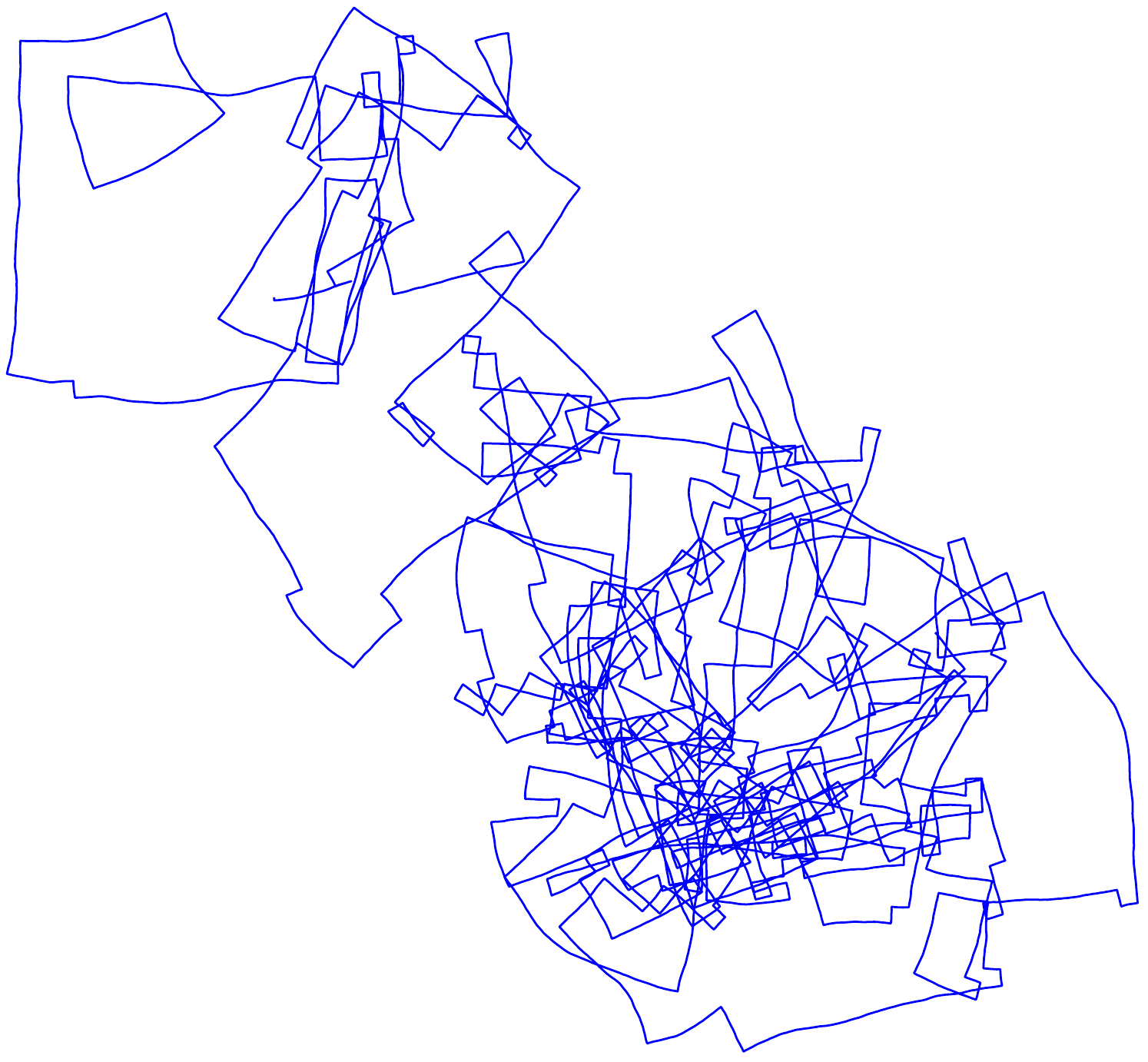}& \includegraphics[width=0.155\textwidth] {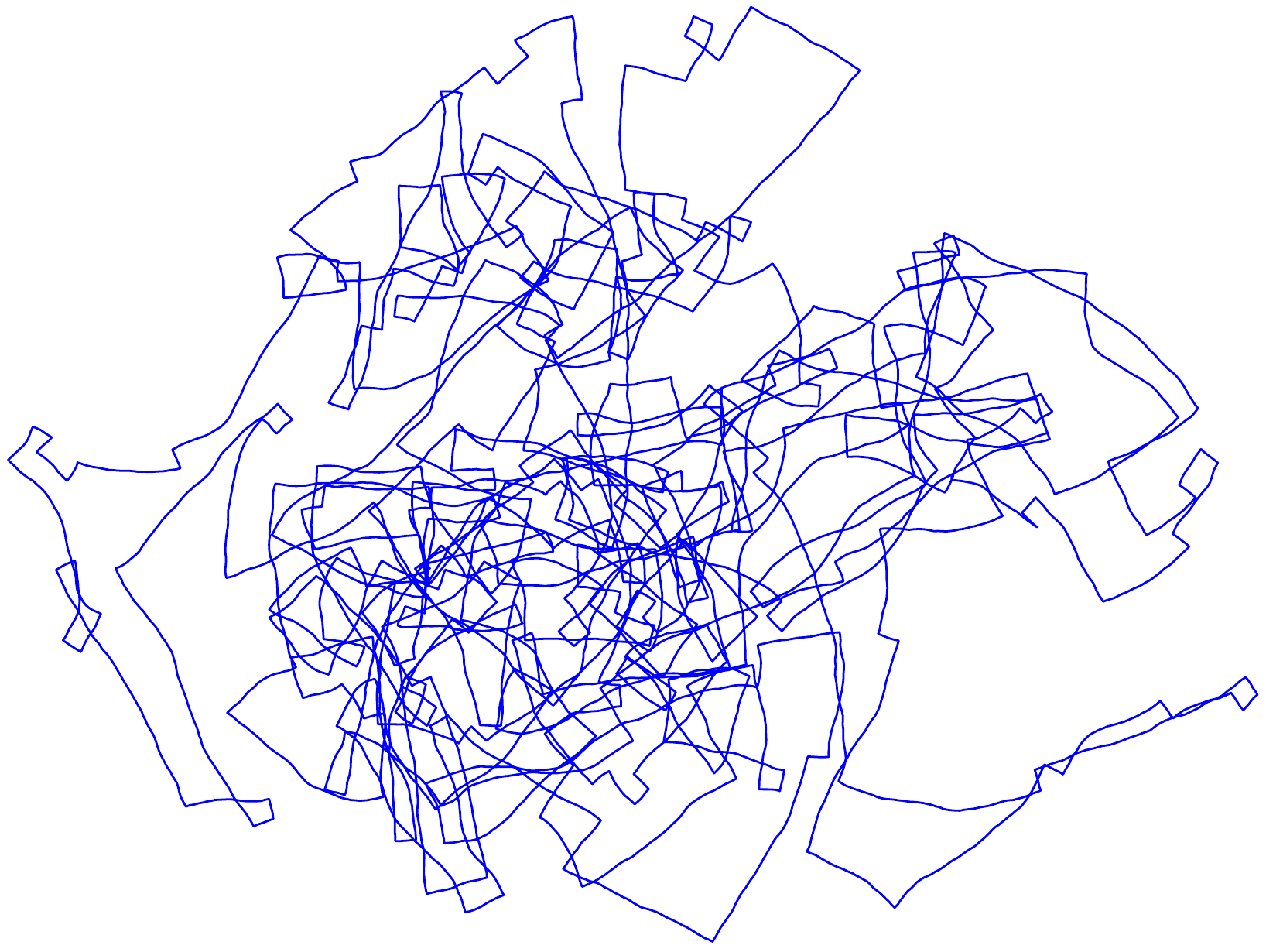}& 
			\includegraphics[width=0.155\textwidth] {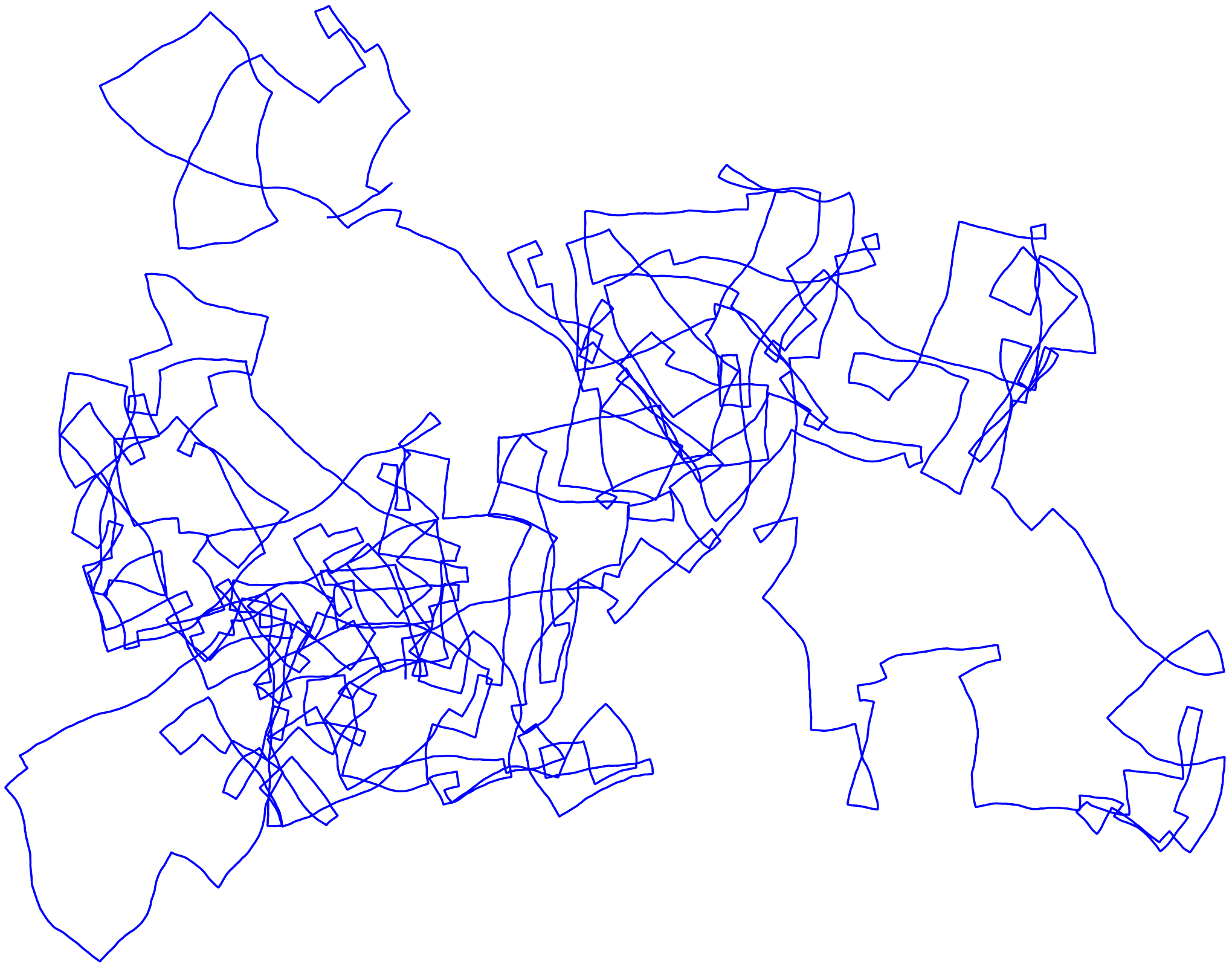}\\
			\etrans{2.58}{0} & \etrans{3.59}{0} & \etrans{2.82}{0} & \etrans{9.77}{-1} & \etrans{1.28}{0} & \etrans{1.36}{0}\\
			\erot{2.96}{1} & \erot{3.99}{1} & \erot{3.87}{1} & \erot{2.53}{0} & \erot{4.84}{0} & \erot{4.73}{0}\\
		\end{tabular}\\
		
		\rotatebox[origin=c]{90}{iSAM} 	
		\begin{tabular}{cccccc}
			\includegraphics[width=0.16\textwidth] {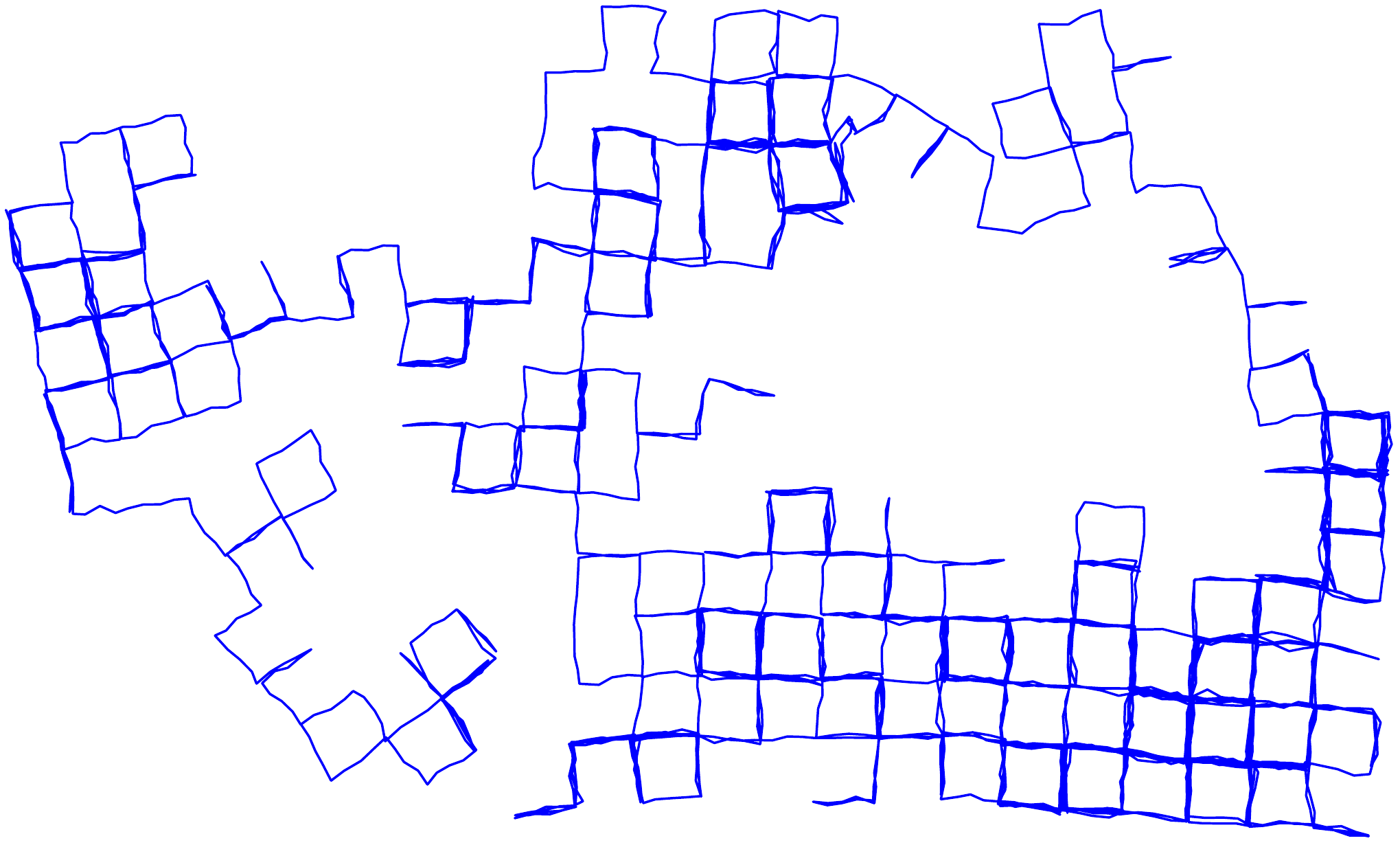} &  \includegraphics[width=0.16\textwidth] {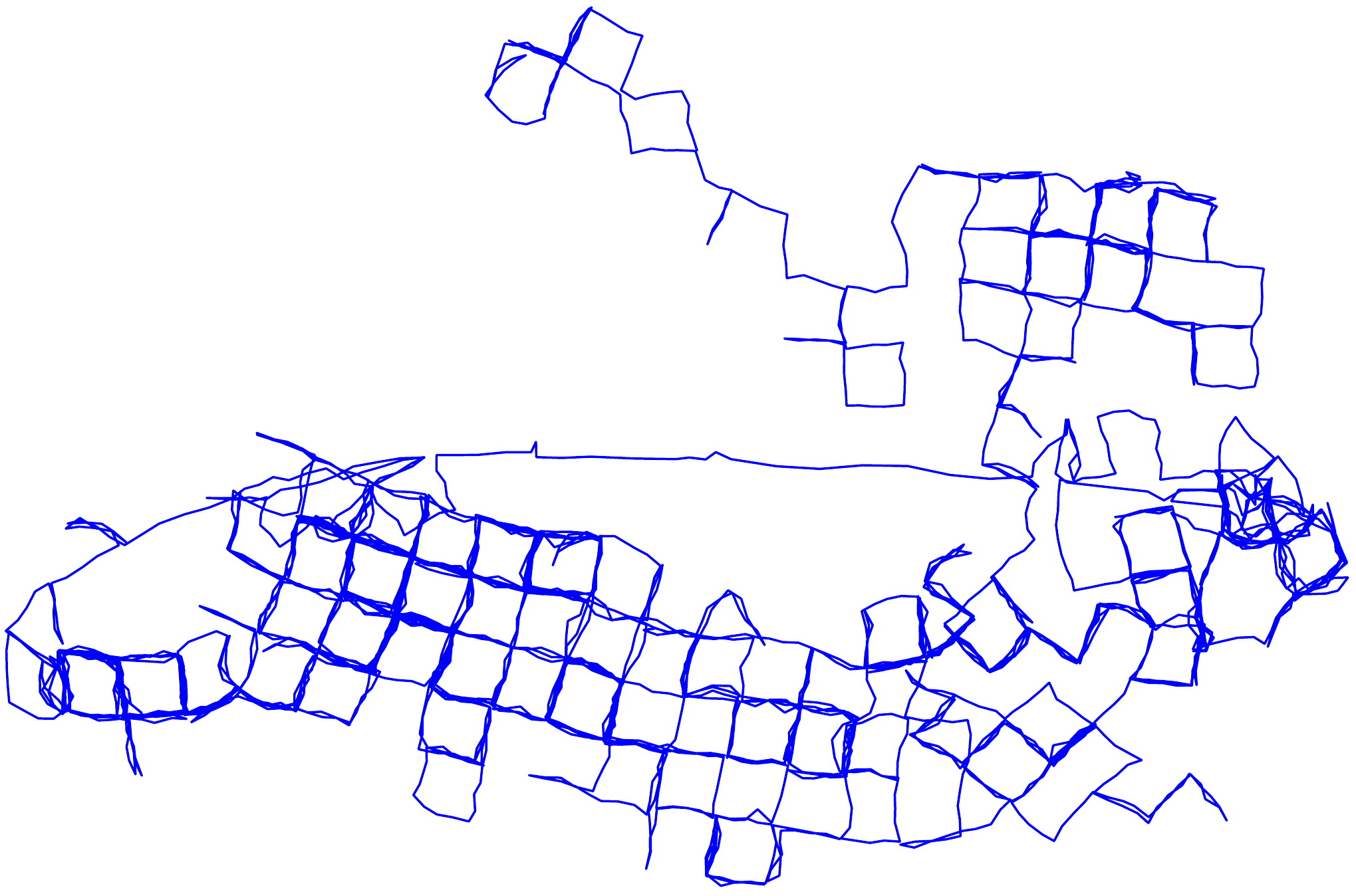} & \includegraphics[height=0.16\textwidth] {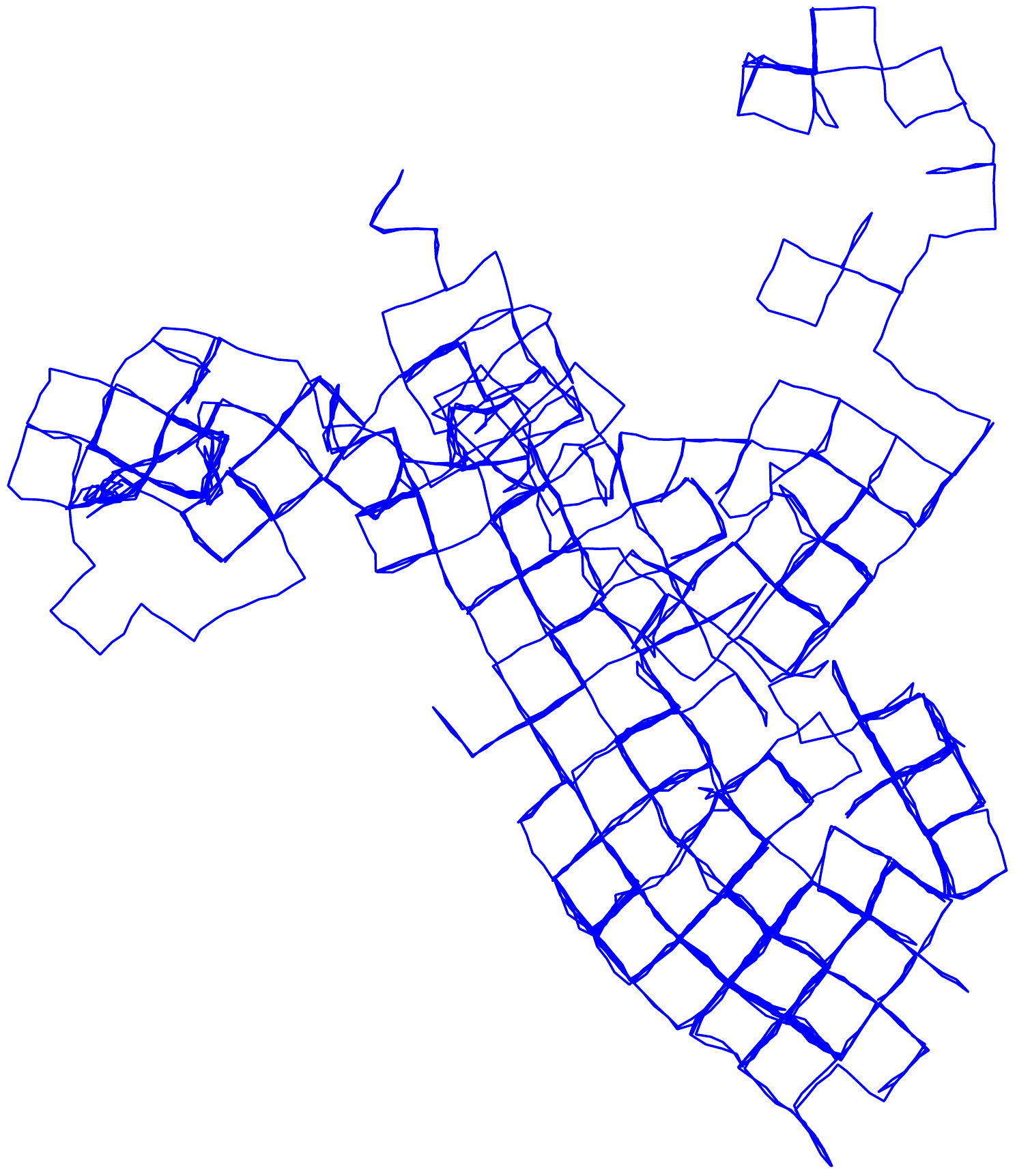} & \includegraphics[width=0.155\textwidth] {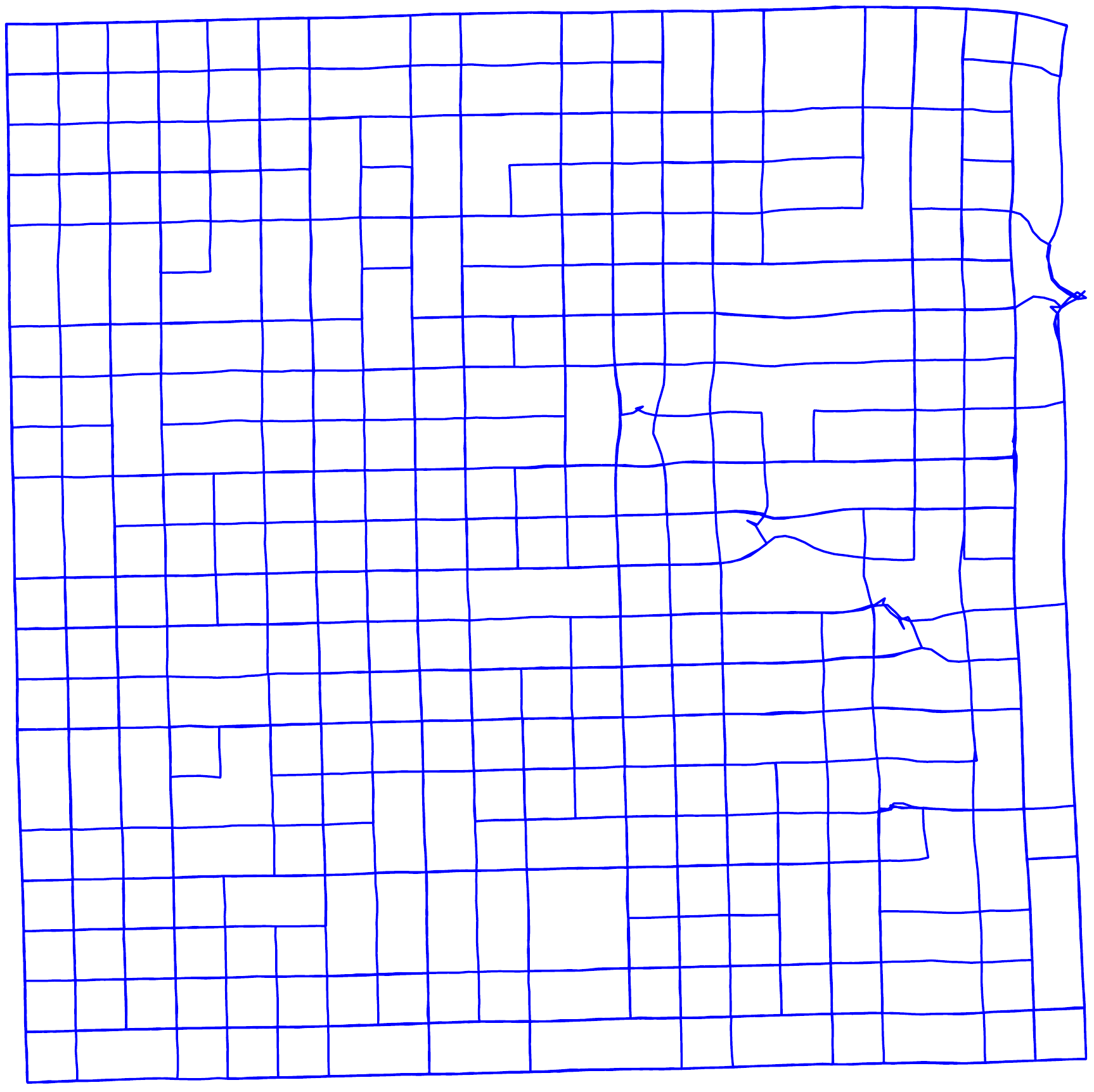} & \includegraphics[width=0.155\textwidth] {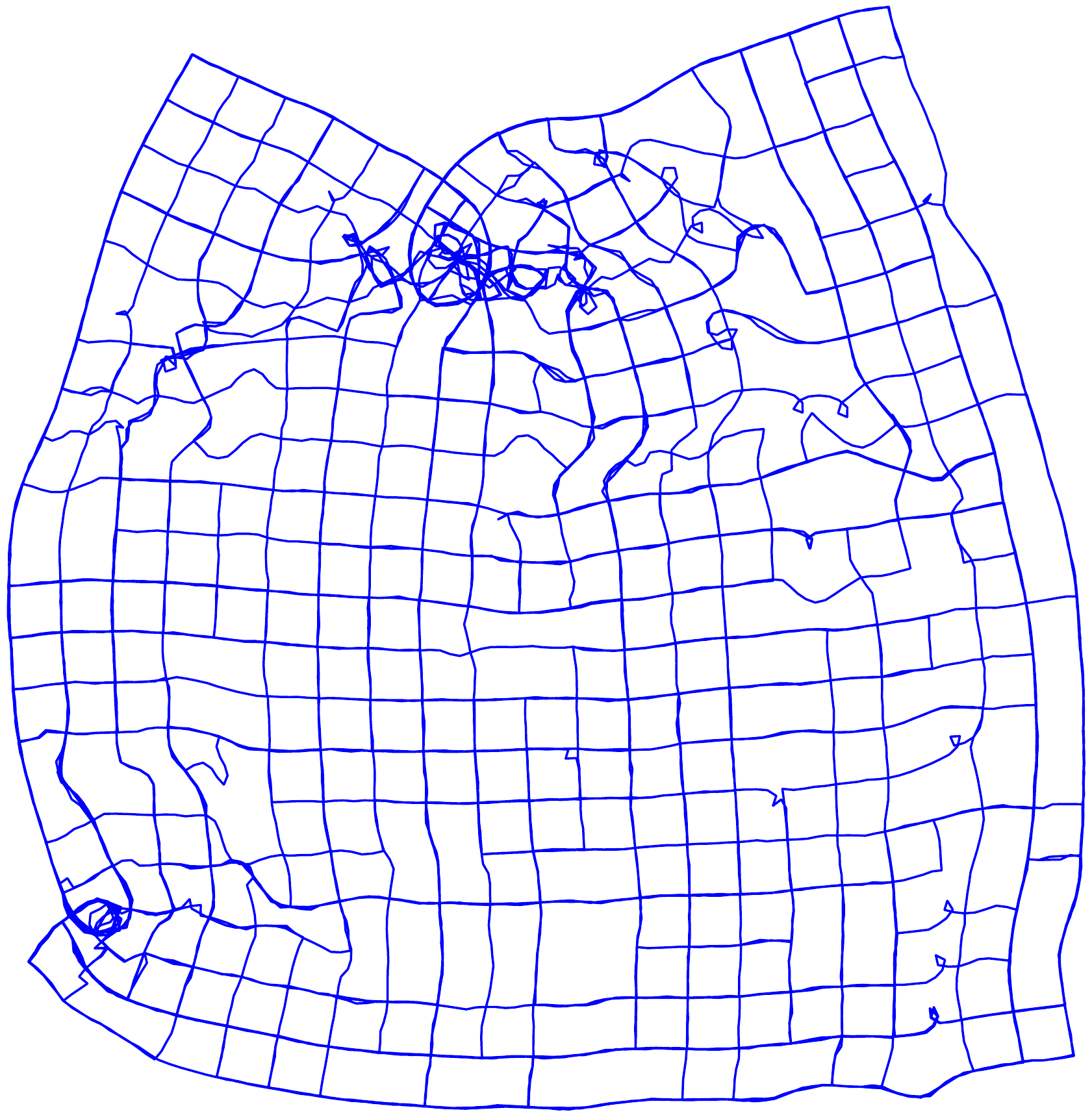} & \includegraphics[width=0.155\textwidth] {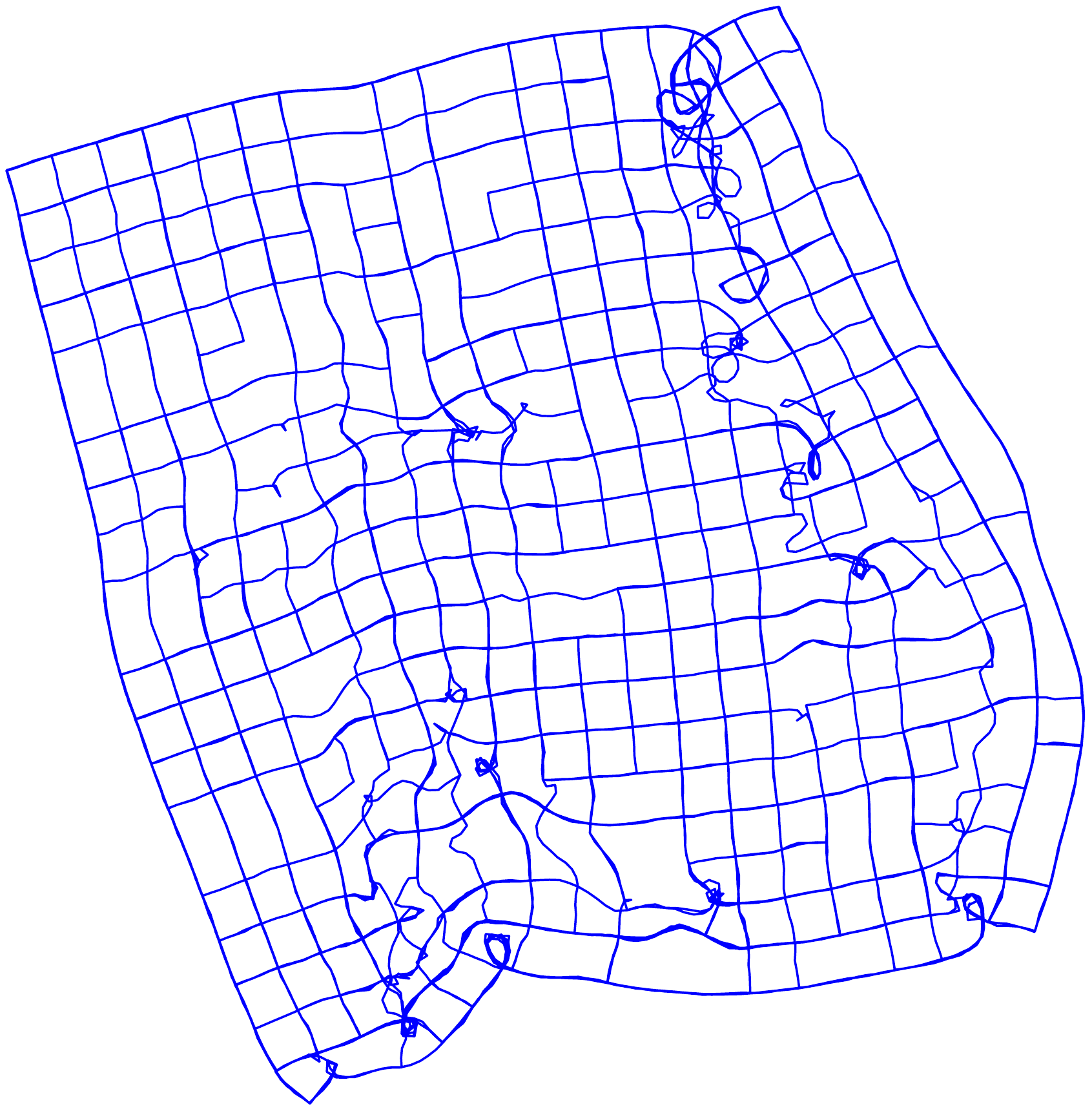}\\
			\etrans{2.80}{-1} & \etrans{1.68}{0} & \etrans{1.27}{0} & \etrans{9.20}{-2} & \etrans{4.85}{-1} & \etrans{4.80}{-1}\\
			\erot{9.09}{0} & \erot{1.97}{1} & \erot{1.69}{1} & \erot{6.24}{0} & \erot{1.61}{1} & \erot{1.69}{1}\\
		\end{tabular}\\
		
		\rotatebox[origin=c]{90}{g2o}
		\begin{tabular}{cccccc}
			\includegraphics[height=0.16\textwidth] {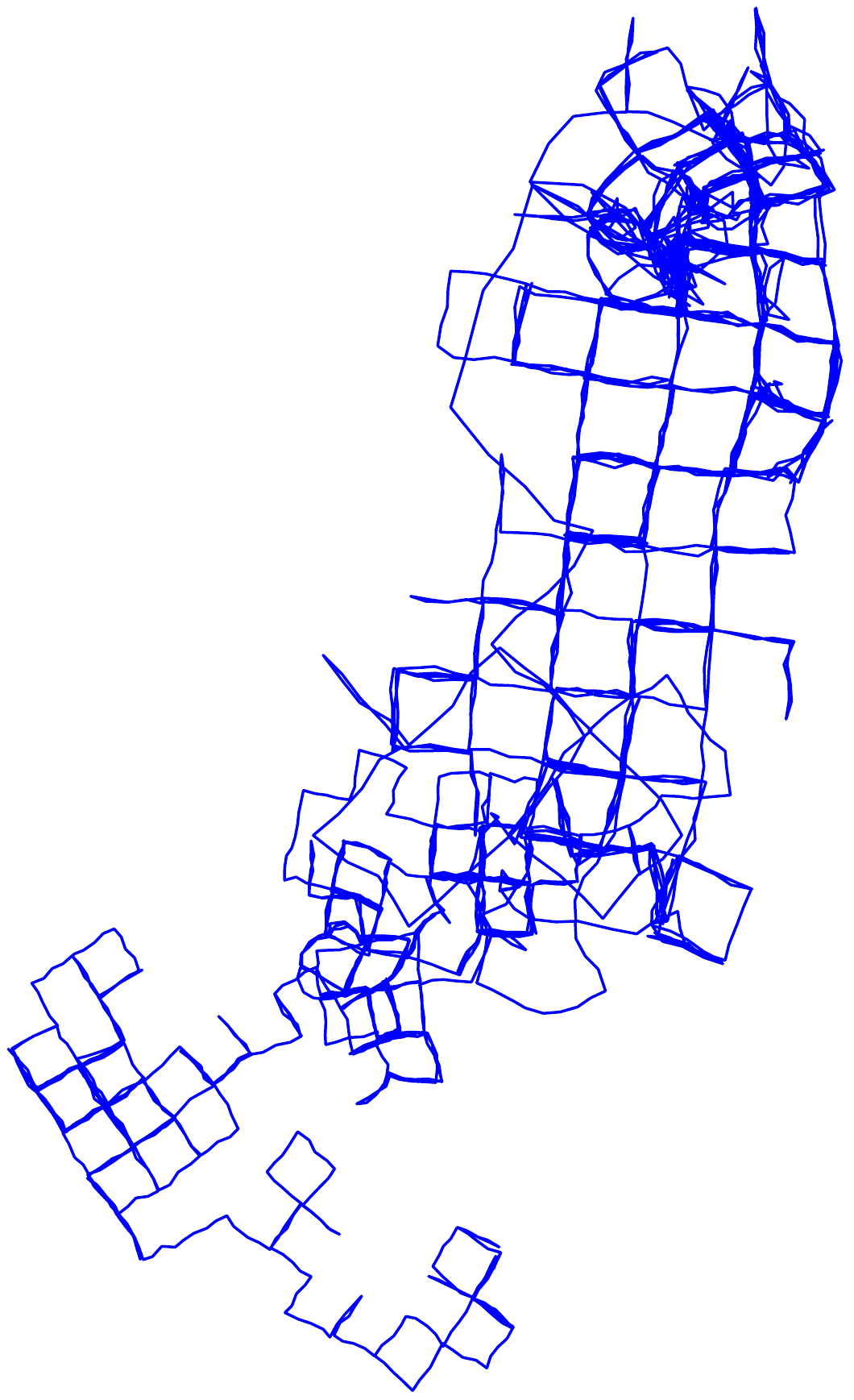}  & \includegraphics[width=0.16\textwidth] {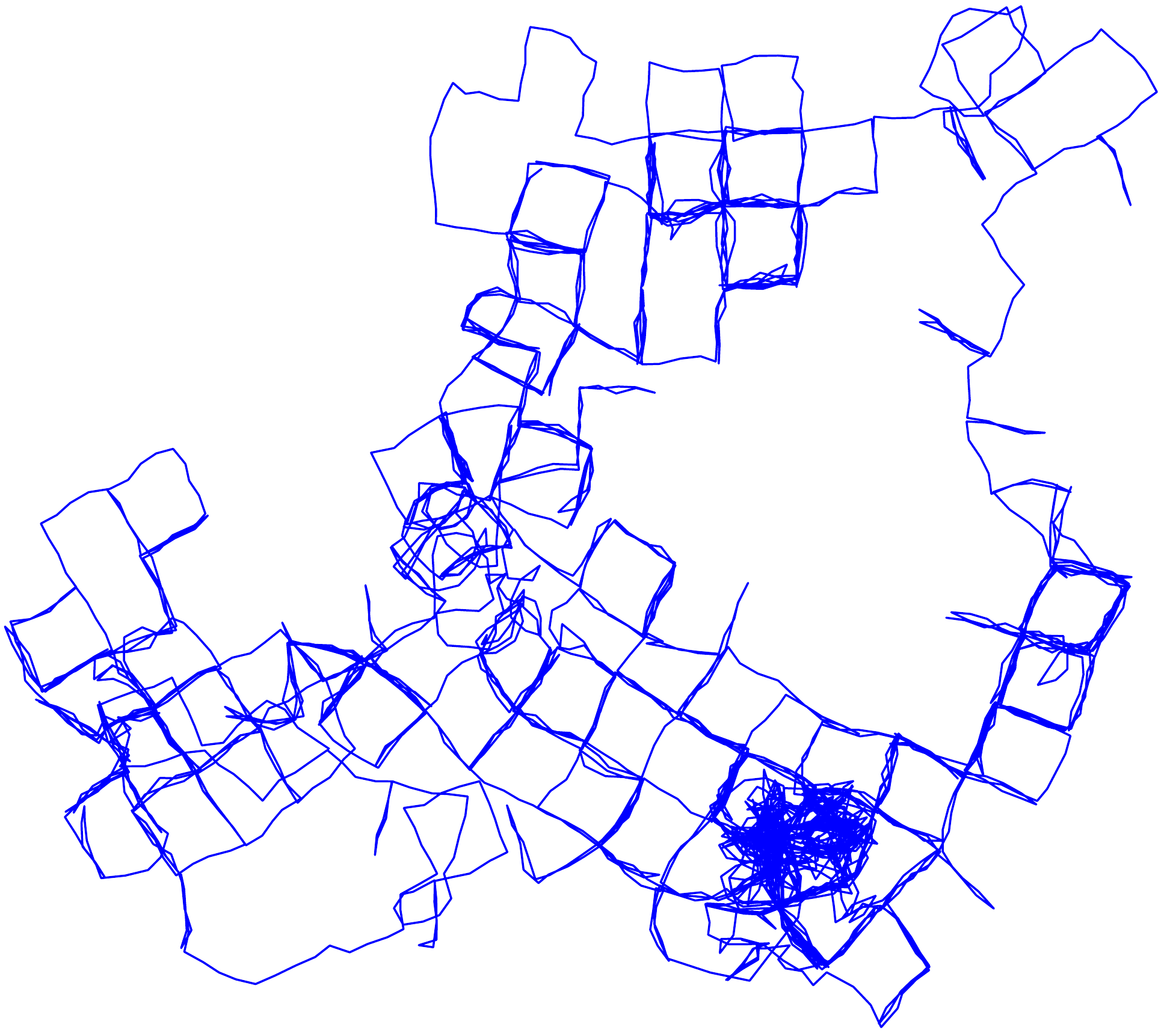} & \includegraphics[width=0.16\textwidth] {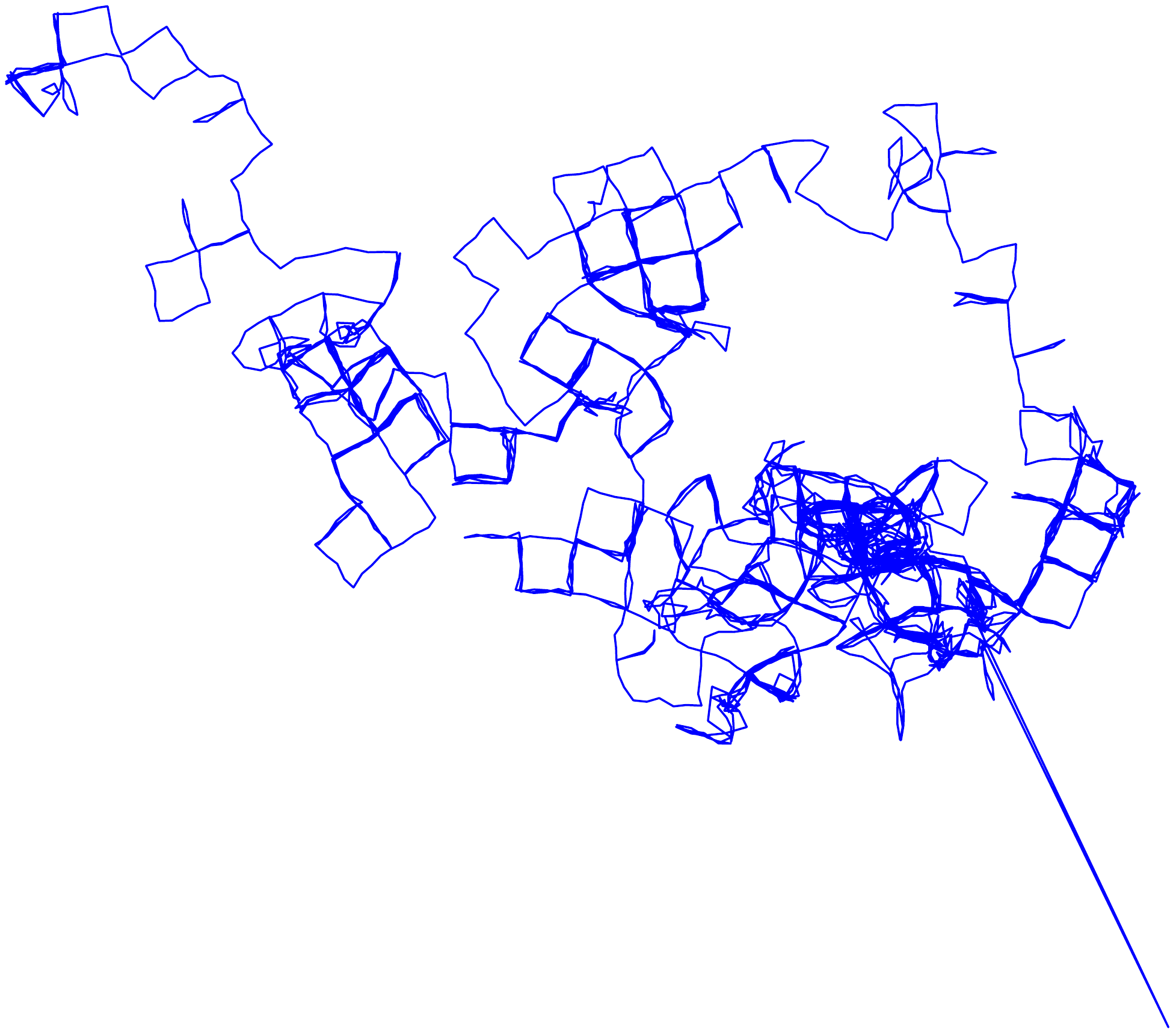} & \includegraphics[width=0.16\textwidth] {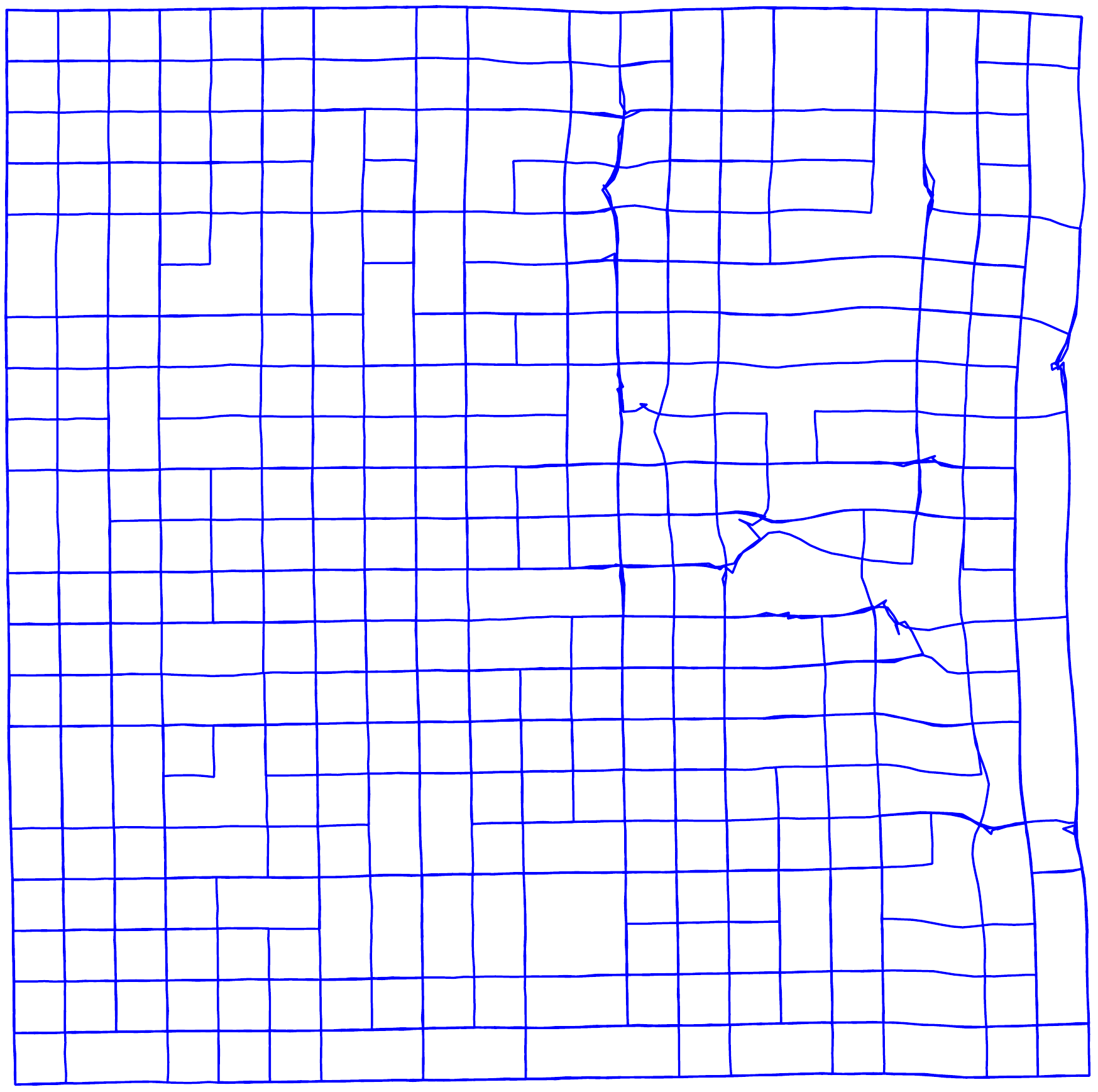} &   \includegraphics[width=0.16\textwidth] {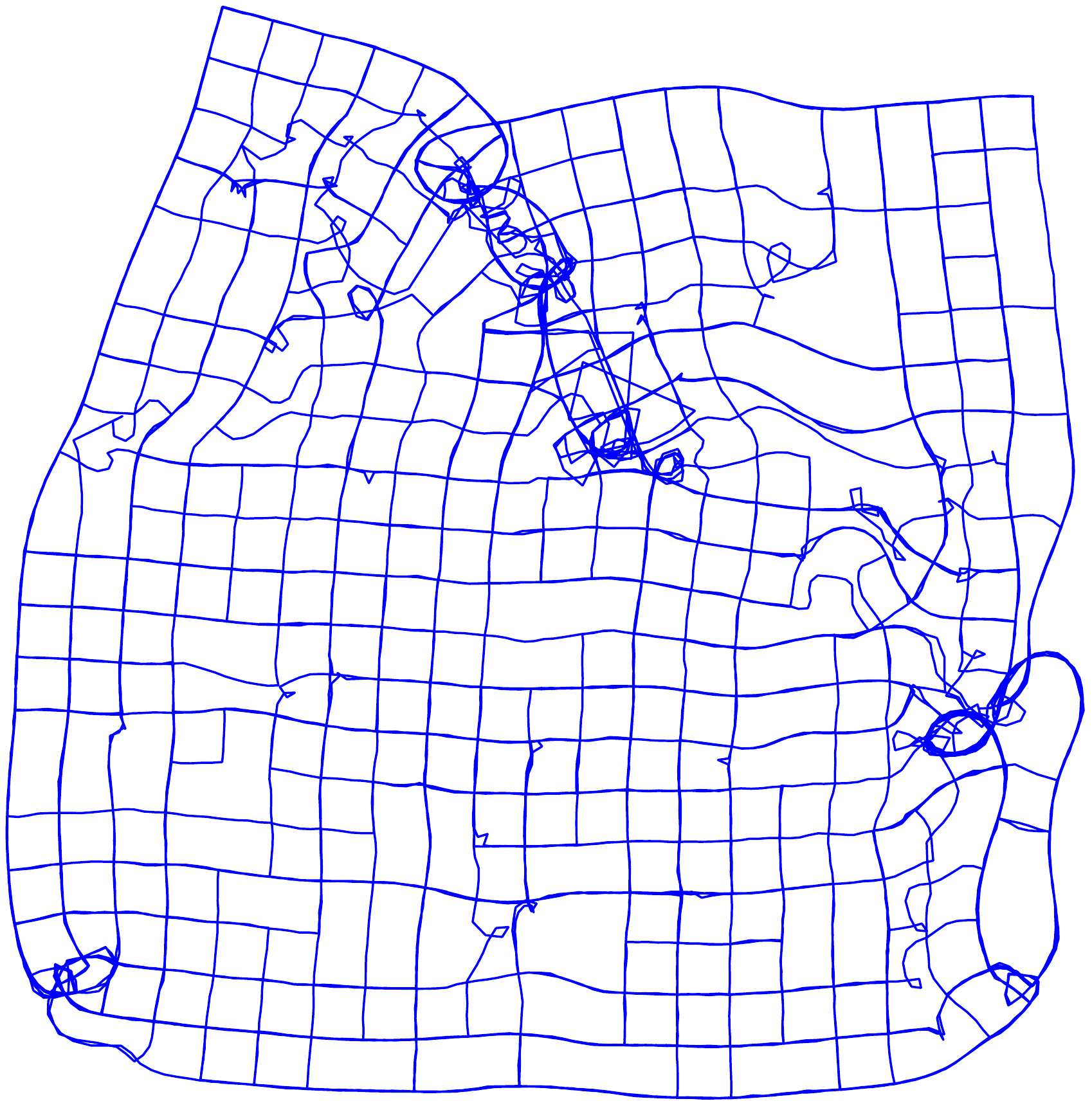} & \includegraphics[width=0.16\textwidth] {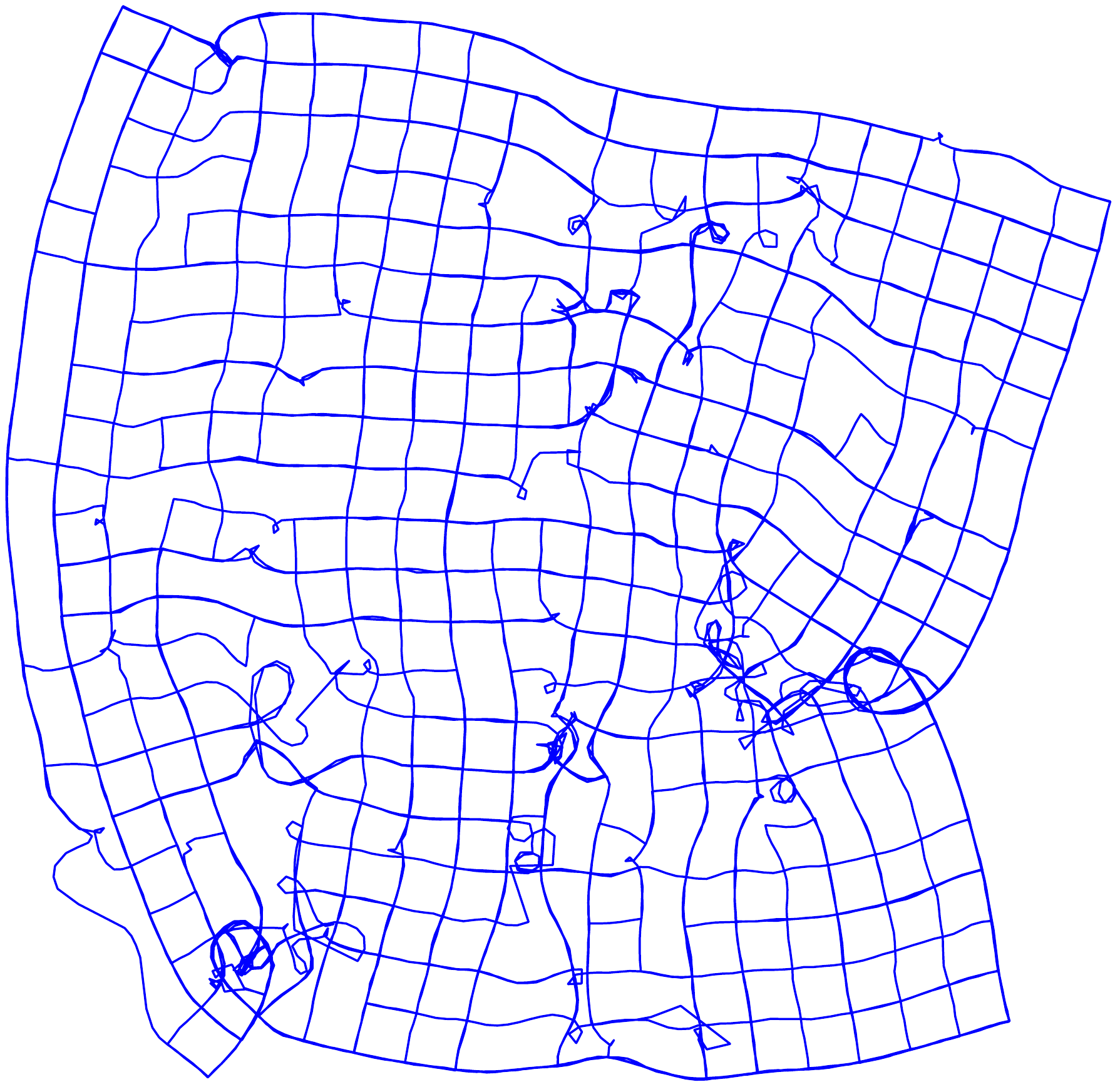}\\
			
			\etrans{1.61}{0} & \etrans{9.47}{-1} & \etrans{1.44}{0} & \etrans{1.07}{-1} & \etrans{6.01}{-1} & \etrans{1.97}{0}\\
			\erot{2.00}{1} & \erot{2.93}{1} & \erot{2.97}{1} & \erot{9.06}{0} & \erot{1.80}{1} & \erot{1.80}{1}\\ 
		\end{tabular}\\
		\rotatebox[origin=c]{90}{RPG-Opt}
		\begin{tabular}{cccccc}
			\includegraphics[width=0.155\textwidth] {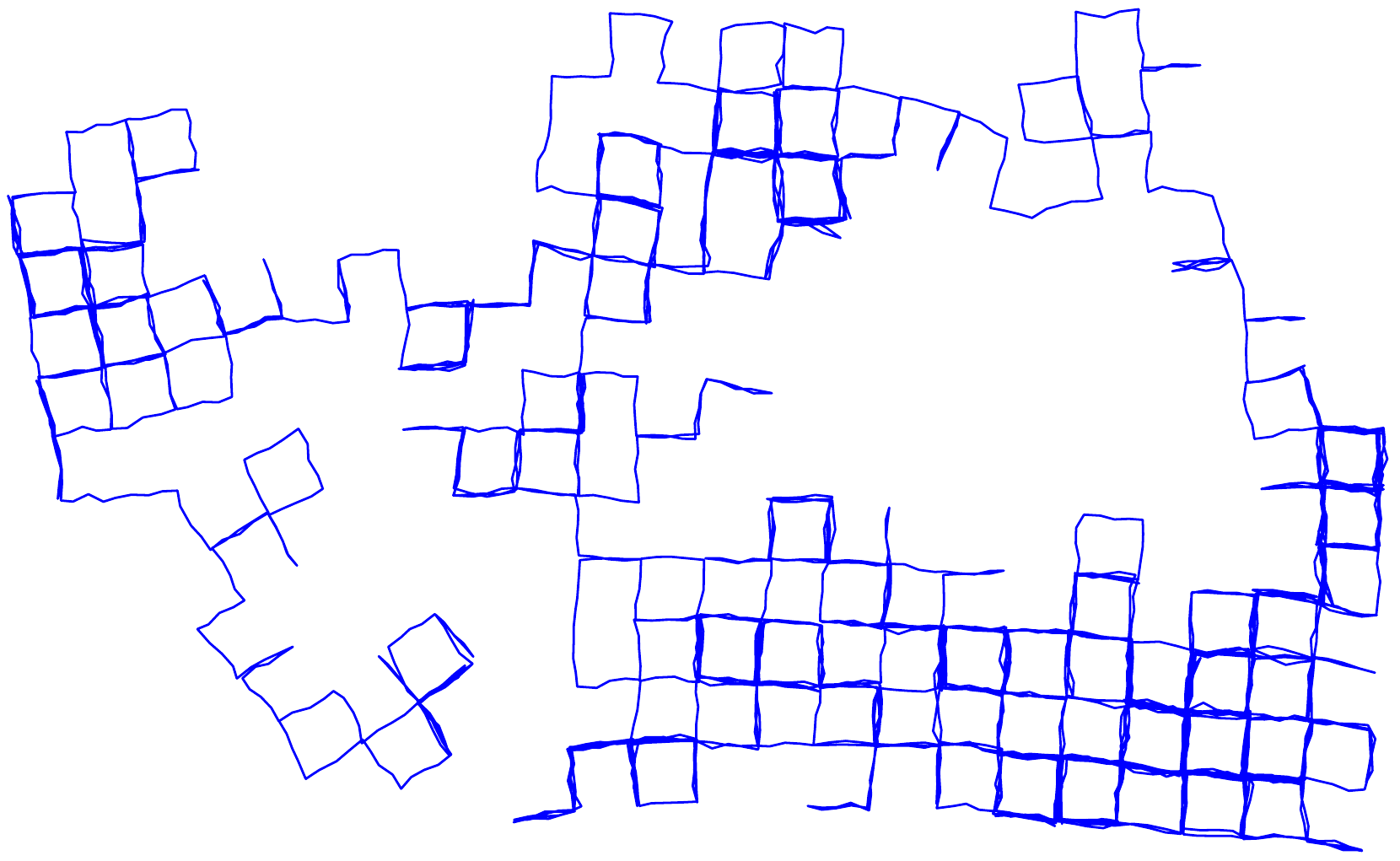} & \includegraphics[width=0.155\textwidth] {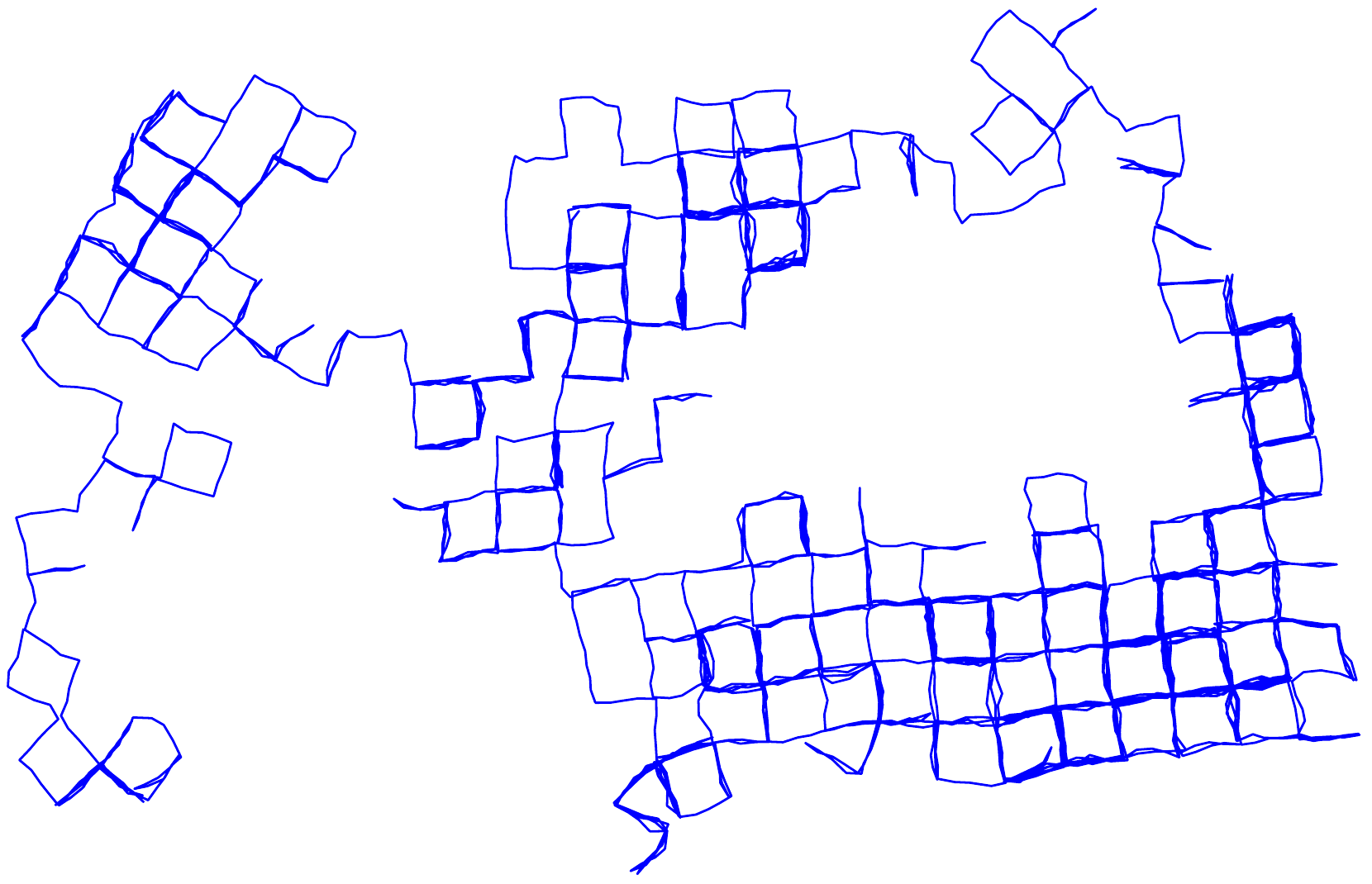} & \includegraphics[width=0.155\textwidth] {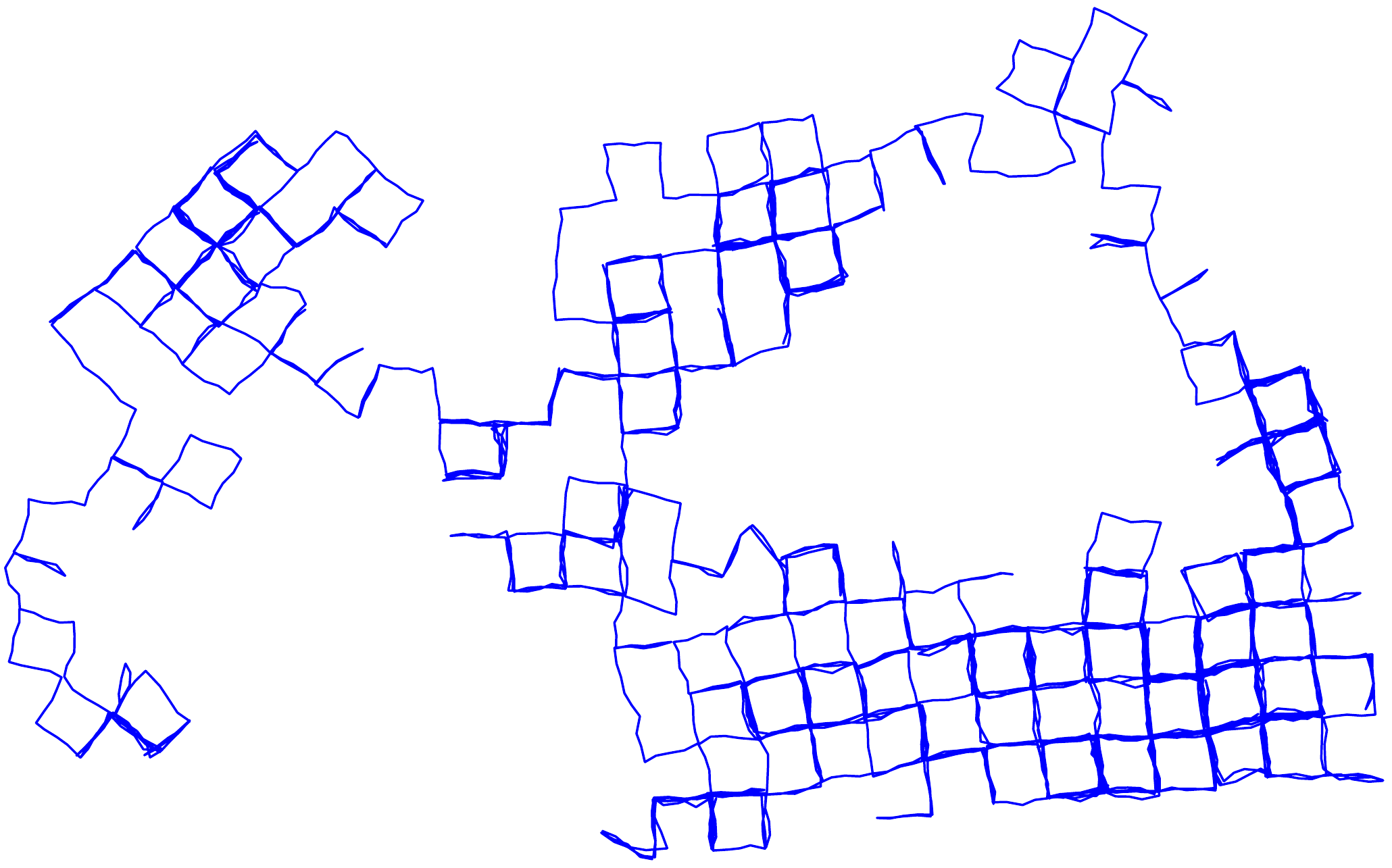} & \includegraphics[width=0.15\textwidth] {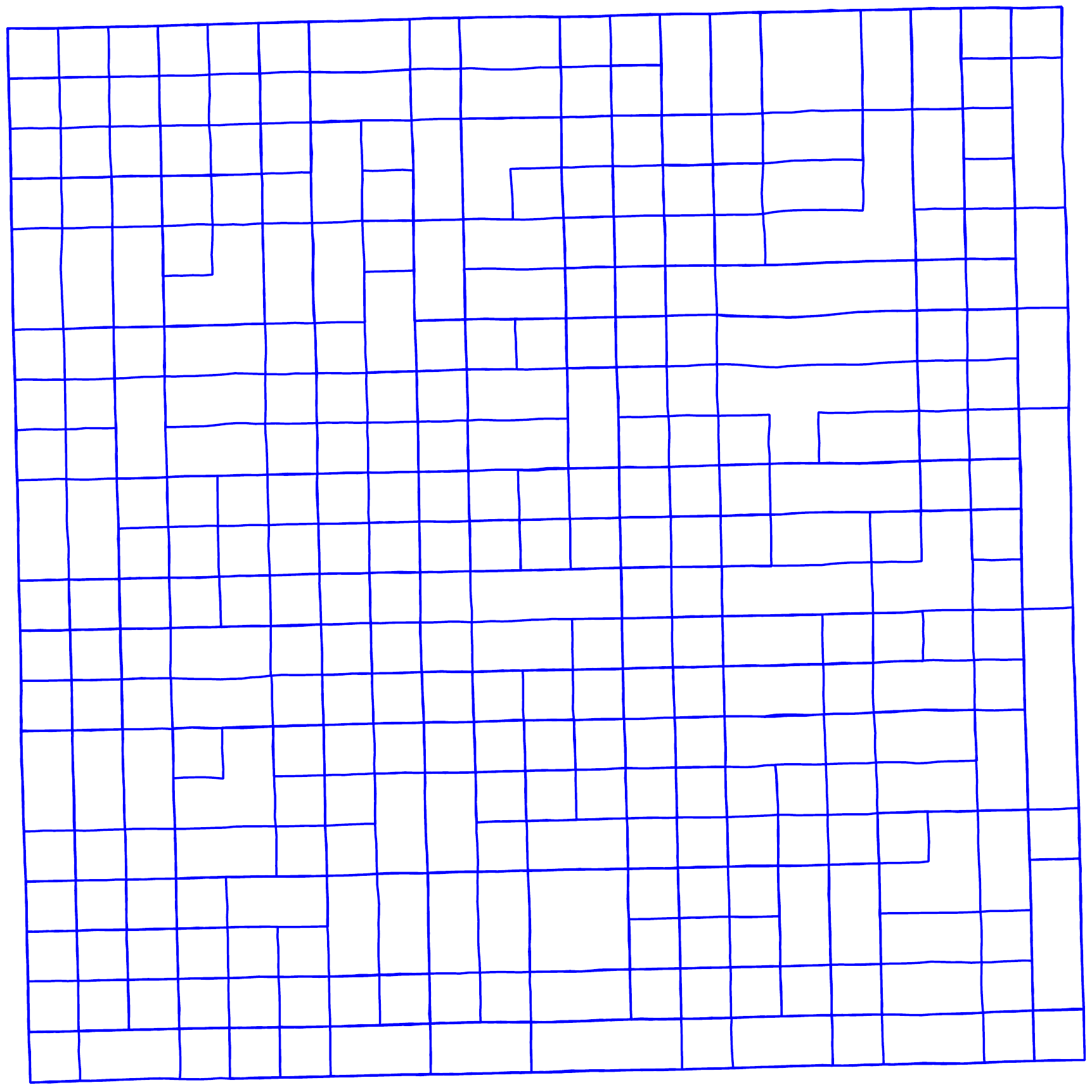} & \includegraphics[width=0.15\textwidth] {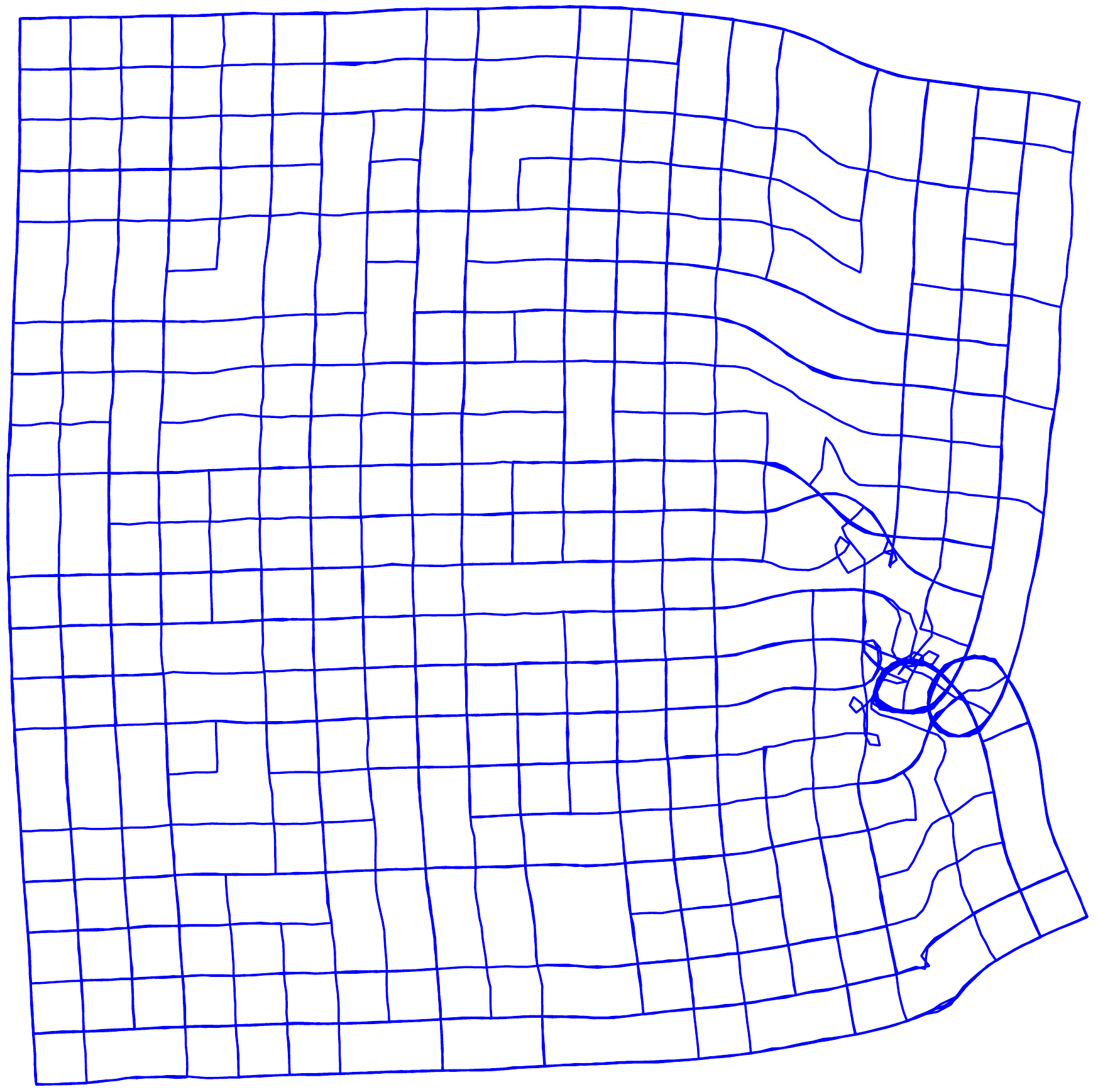} &  \includegraphics[width=0.15\textwidth] {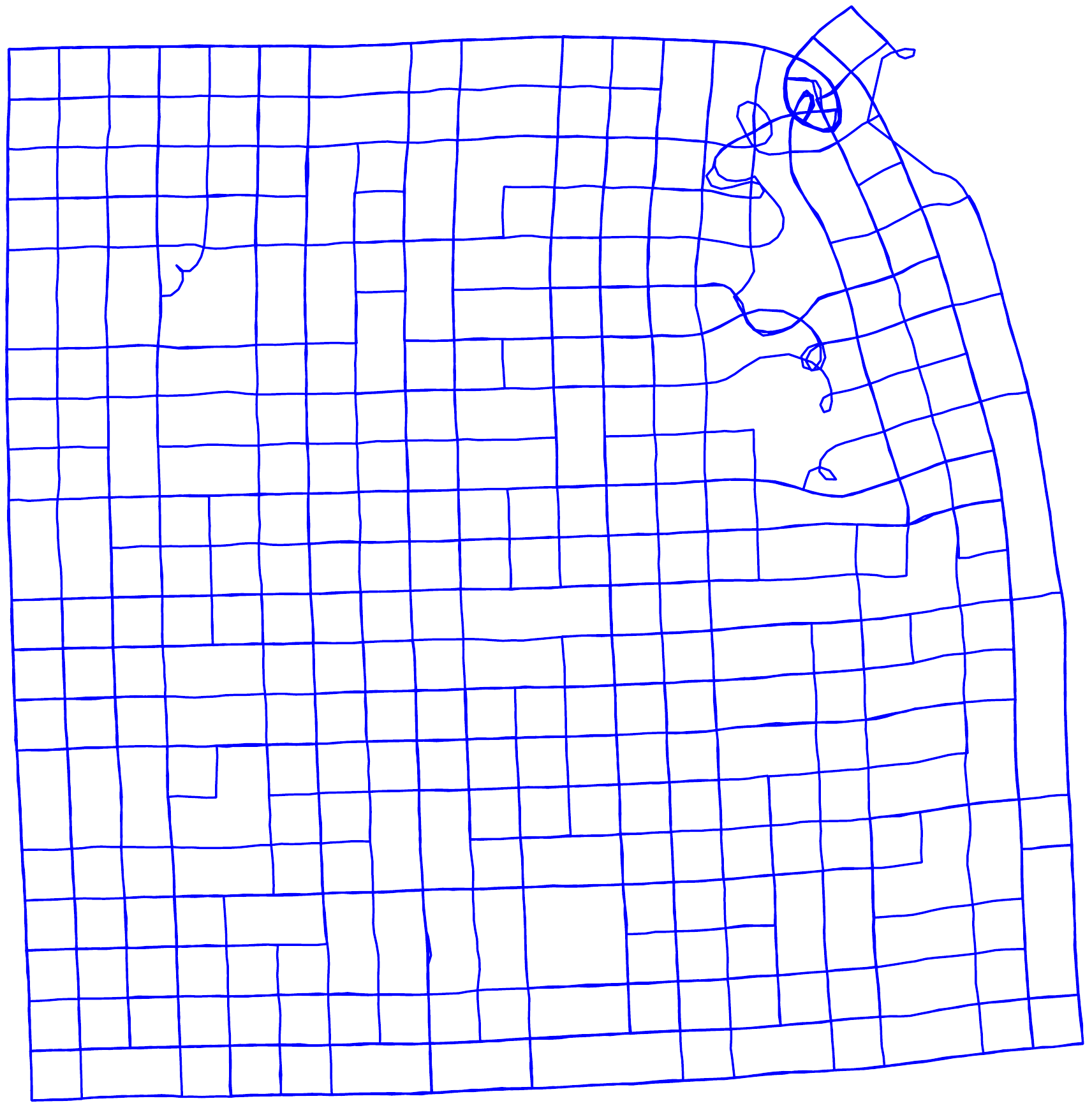} \\
			\etrans{2.60}{-1} & \etrans{3.79}{-1} & \etrans{3.94}{-1} & \etrans{4.34}{-2} & \etrans{3.20}{-1} & \etrans{2.86}{-1}\\
			\erot{8.42}{0} & \erot{1.31}{1} & \erot{1.31}{1} & \erot{1.55}{0} & \erot{6.95}{0} & \erot{5.79}{0}\\
		\end{tabular}
		\caption{Evaluation results using synthetic data sets. Here, the optimizations are initialized with odometry measurements, and iteration steps for RPG-Opt and g2o are fixed to be $30$\,.  Results from GTSAM are not listed because of nonconvergence. The frameworks iSAM and g2o are prone to local minima. Though initialized directly with the odometry measurements, the proposed RPG-Opt shows the best accuracy and robustness against local minima under large uncertainty. Data sets with `a' (e.g., \textit{M3500a+} or \textit{City10000a}) are added with equivalently scaled odometry noise for rotations and translations. Data sets with `b' have unequal noise level of rotation and translation.  Data sets with `c' have unequal and correlated translational and rotational noise. The RPEs for rotation are in angular degree. The additional odometry noise is set up given the covariance $\bm{\Sigma}$, thus $\bm{\Omega}_{ij}=\bm{\Sigma}^{-1}$\,.}
		\label{fig:hard}
	\end{figure*}
	
	The retraction maps a point from the tangent plane back to the manifold, namely $\mathcal{R}_{\uvx}:\mathds{T}_{\uvx}\mathbb{M}^n\rightarrow\mathbb{M}^n$\,, which is not unique. For instance, there have been different projection-like retractions proposed in~\cite{absil2012projection}. We use the exponential retraction for $\uvx$ with each node undergoing the exponential map defined in (\ref{eq:exp}) for update, namely
	\begin{equation*}
	\vx_i^{k+1}=\Exp_{\vx_i^k}(\valpha_i^k)\,,
	\end{equation*}
	for $i = 1,\,...\,,n$\,. After the retraction back to the manifold, the cost function can then be linearized at the tangent plane at $\tilde{\vx}^{k+1}$. The optimization is stopped until the norm of the Riemannian gradient becomes sufficiently small. 
	\begin{figure}[t]
		\centering
		\newcommand{\erot}[2]{$e_\text{r} = #1 \cdot 10^{#2}$}
		\newcommand{\etrans}[2]{$e_\text{t} = #1 \cdot 10^{#2}$}
		$\bm{\Sigma}$
		\begin{tabular}{cc}
			\textit{M3500d+} & \textit{City10000d} \\	
			$\Big[\sbbmat 0.04&0.015&0.03\\0.015&0.04&0.03\\0.03&0.03&0.13\sebmat\Big]$ & $\Big[\sbbmat 0.04&0.015&0.03\\0.015&0.04&0.03\\0.03&0.03&0.13\sebmat\Big]$ \\	
		\end{tabular}\\
		\rotatebox[origin=c]{90}{chord}
		\begin{tabular}{cc}
			s\includegraphics[width=0.16\textwidth] {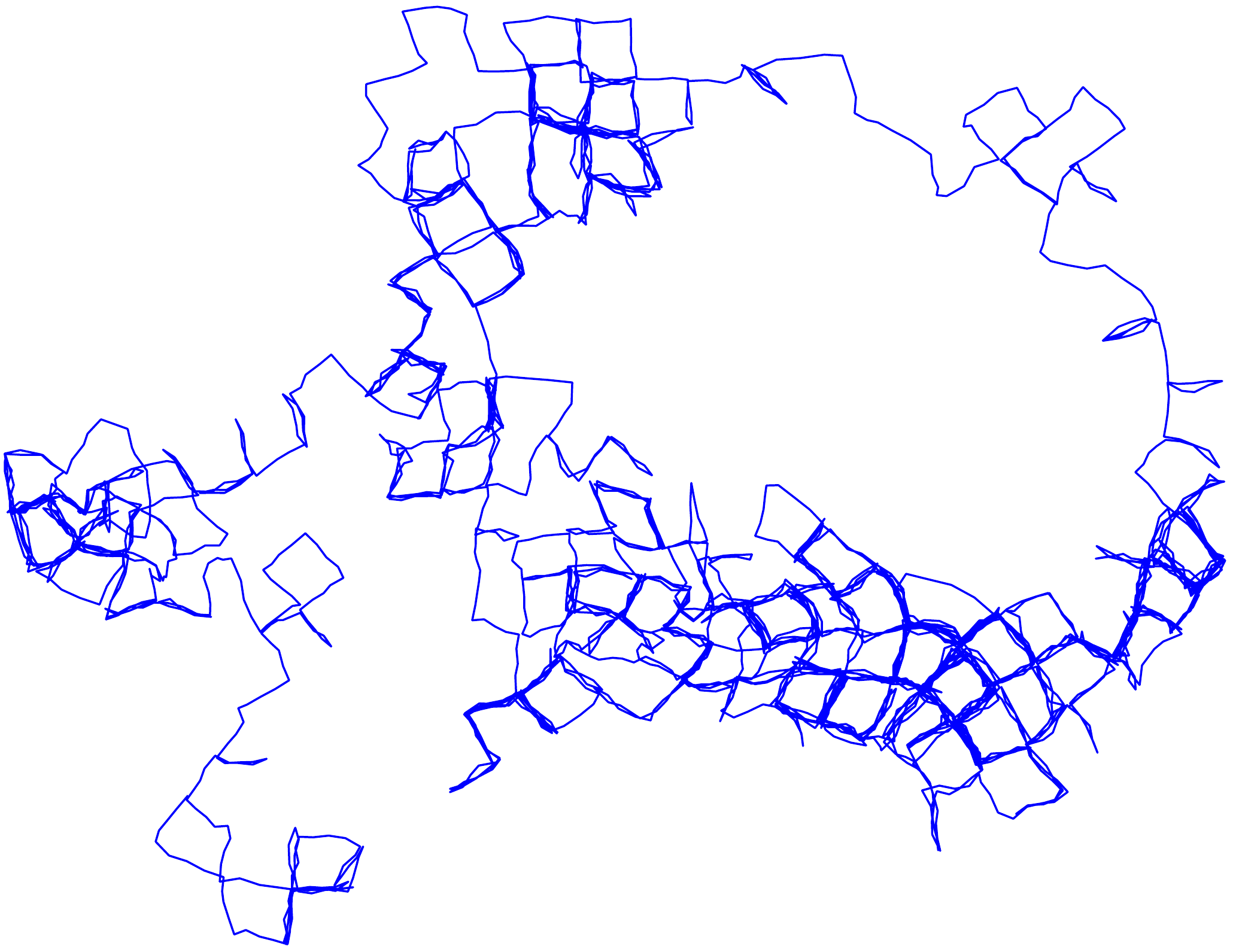} & \includegraphics[width=0.16\textwidth] {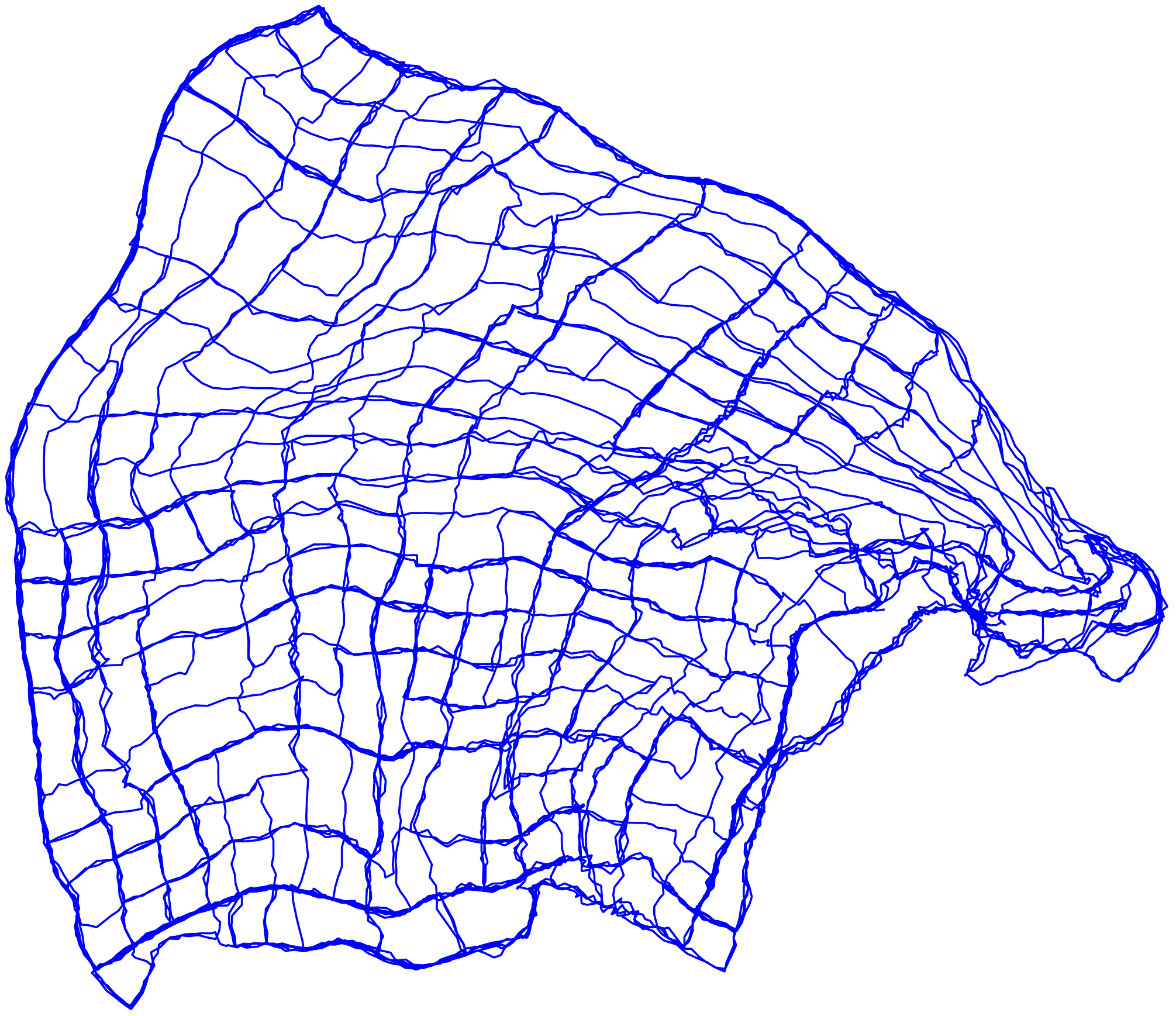} \\
			\etrans{7.08}{-1} & \etrans{7.23}{-1}  \\
			\erot{1.85}{1} & \erot{1.64}{1} \\
		\end{tabular}\\
		\rotatebox[origin=c]{90}{chord + GTSAM}
		\begin{tabular}{cc}
			\includegraphics[width=0.16\textwidth]{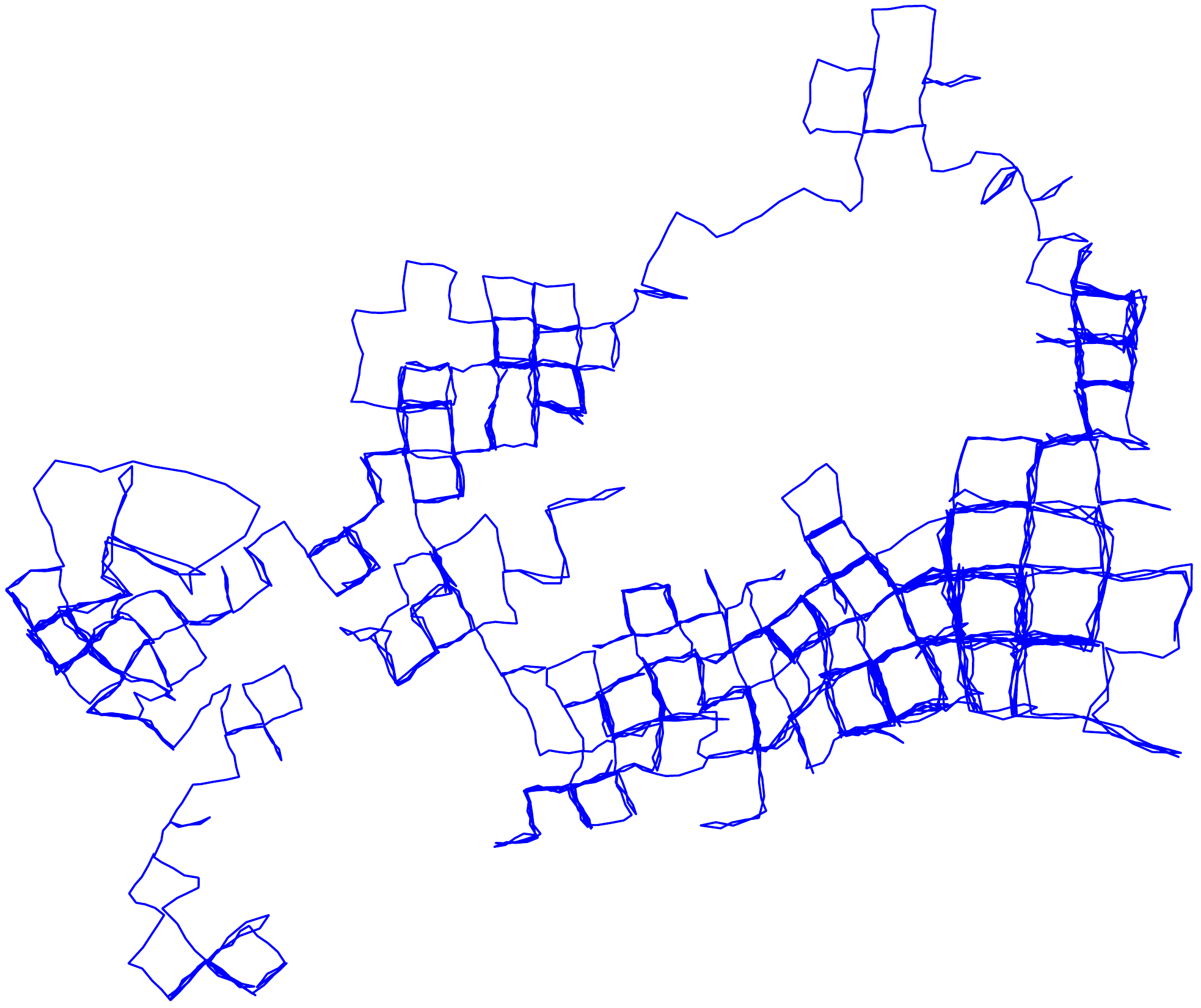} & \includegraphics[width=0.16\textwidth]{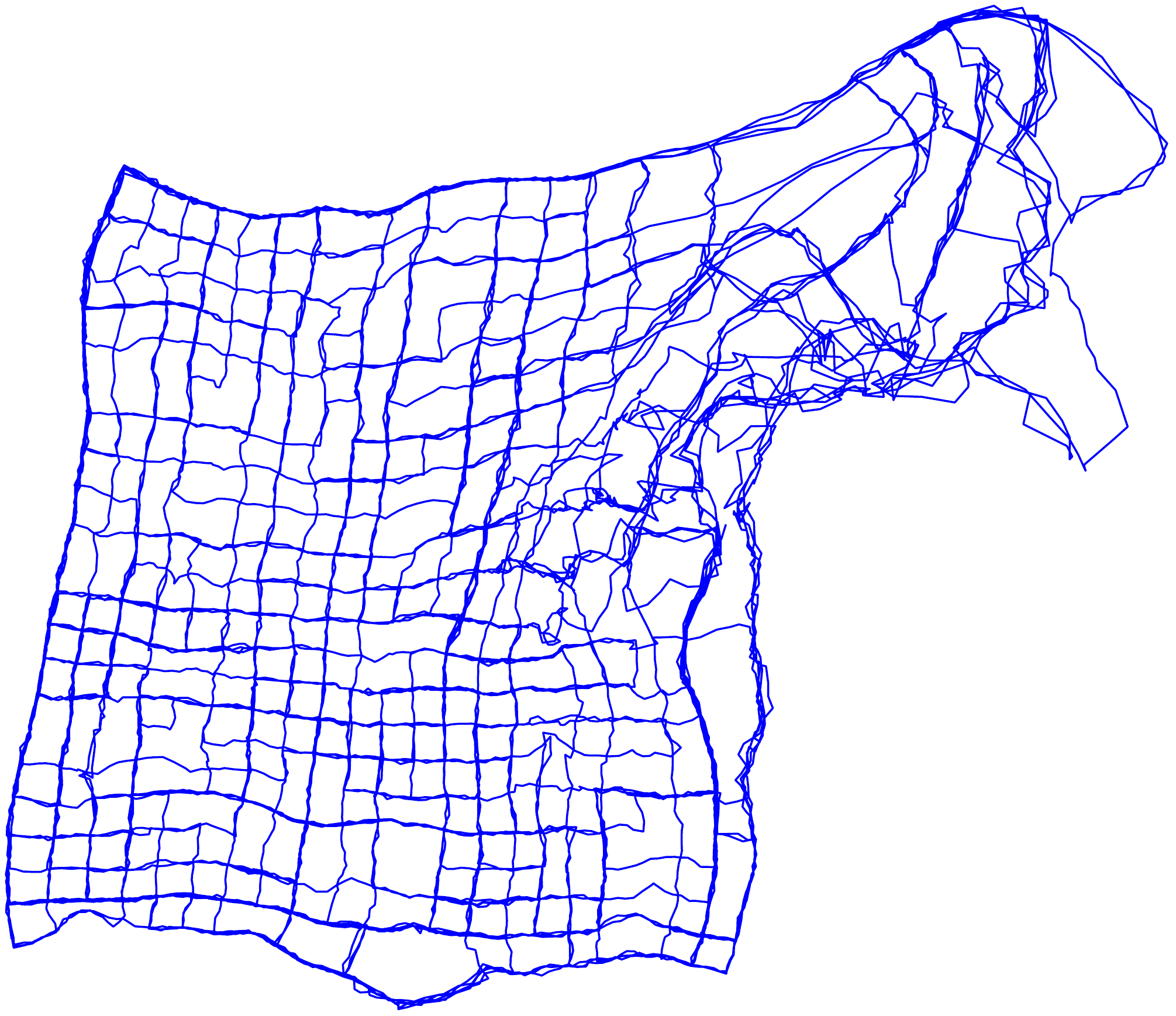} \\
			\etrans{7.01}{-1} & \etrans{9.28}{-1} \\
			\erot{1.57}{1} & \erot{1.39}{1} \\
		\end{tabular}\\
		\rotatebox[origin=c]{90}{chord + g2o} 
		\begin{tabular}{cc}
			\includegraphics[width=0.16\textwidth] {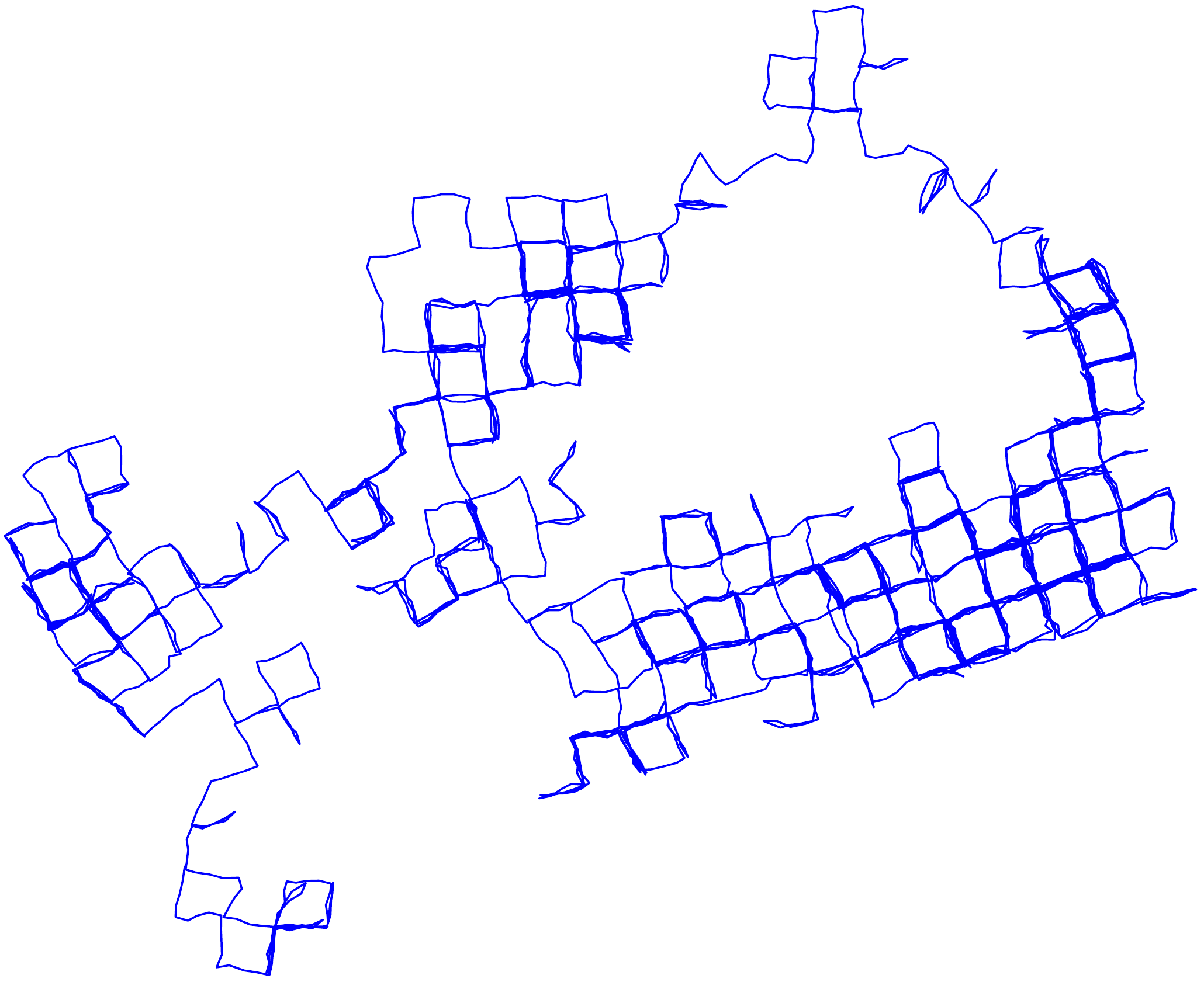} & \includegraphics[width=0.16\textwidth] {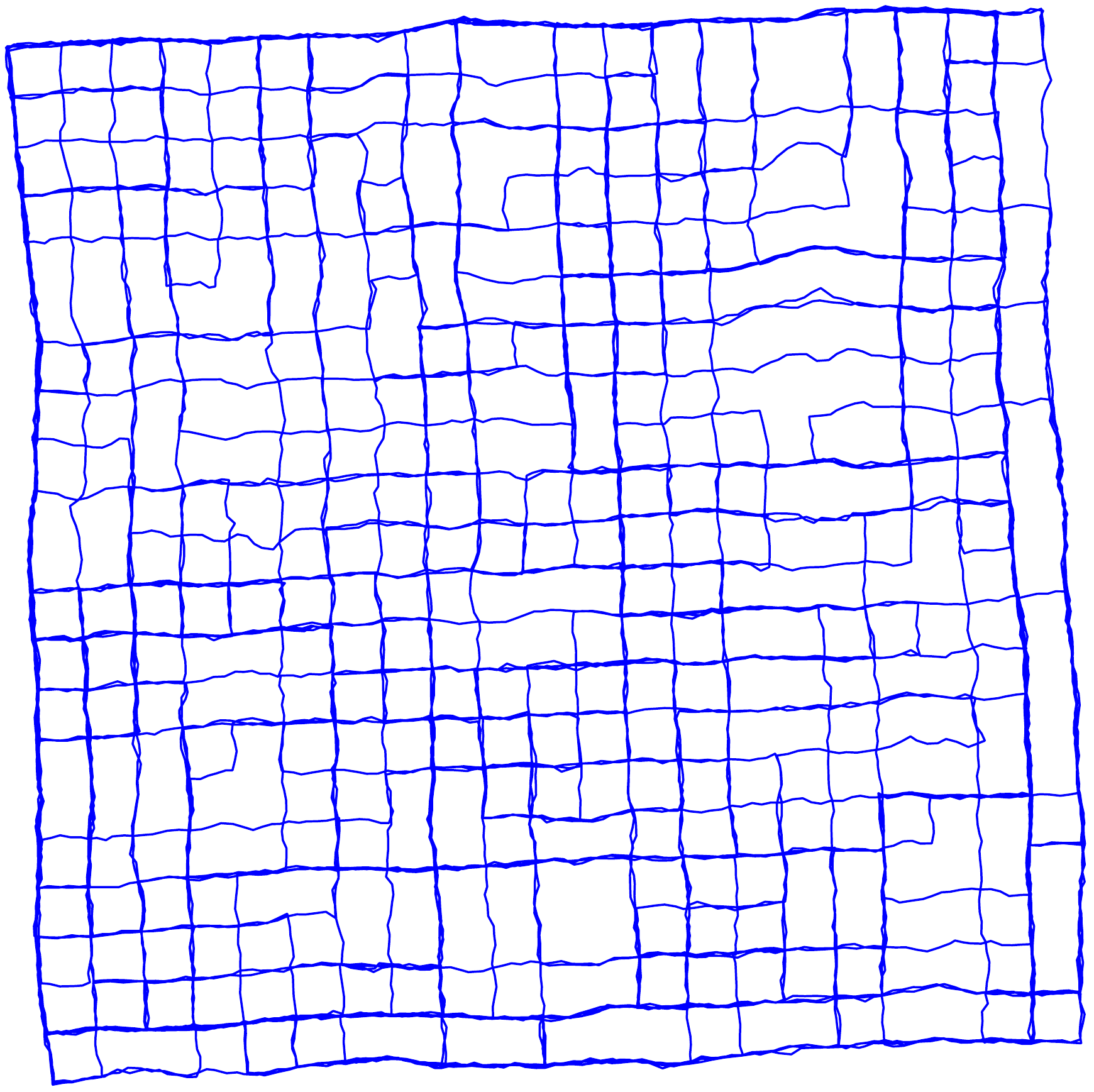} \\
			\etrans{4.78}{-1} & \etrans{2.46}{-1} \\
			\erot{1.51}{1} & \erot{1.23}{1} \\
		\end{tabular}\\
		\rotatebox[origin=c]{90}{chord + RPG-Opt}
		\begin{tabular}{cc}
			\includegraphics[width=0.16\textwidth] {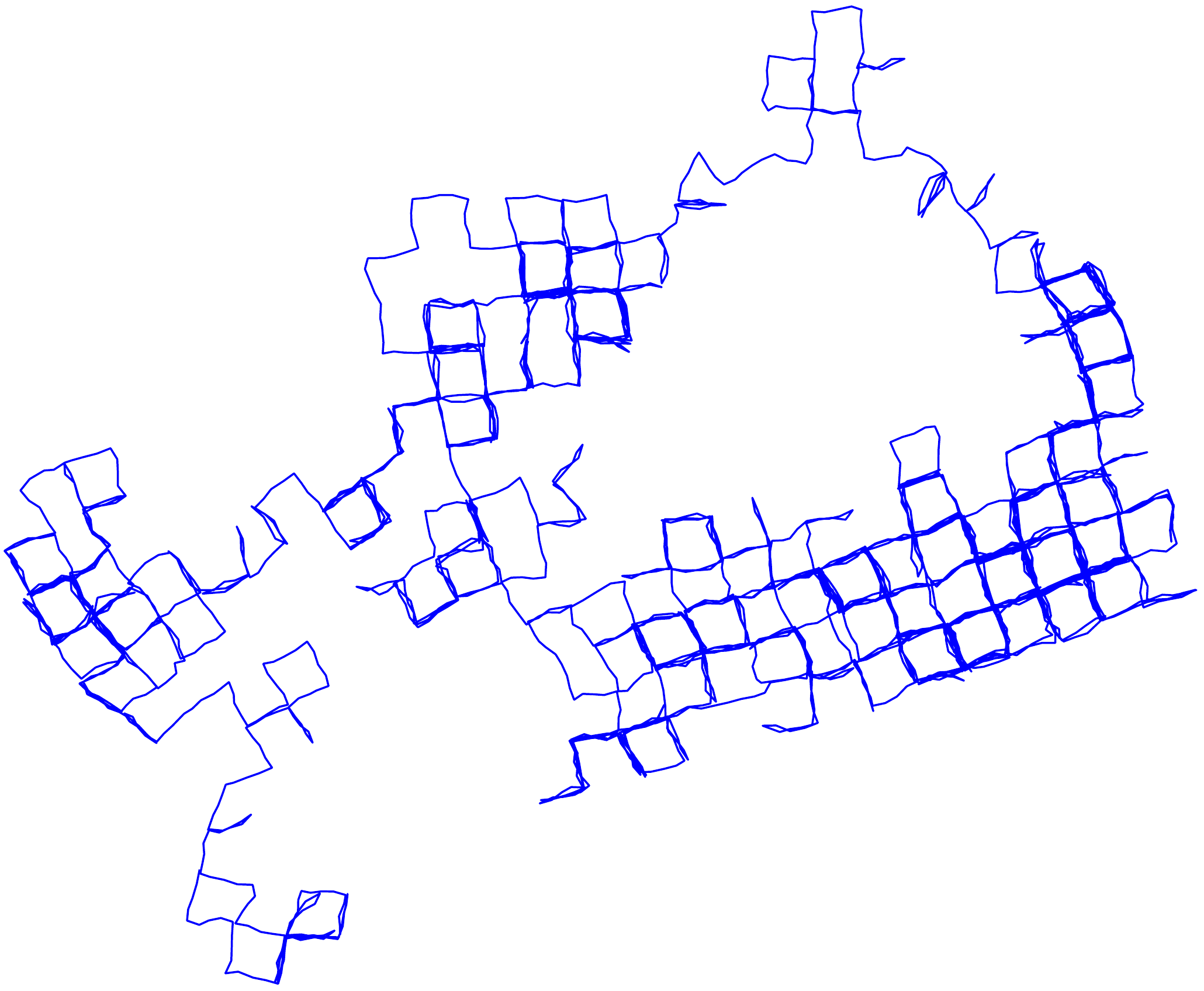}  &  \includegraphics[width=0.16\textwidth] {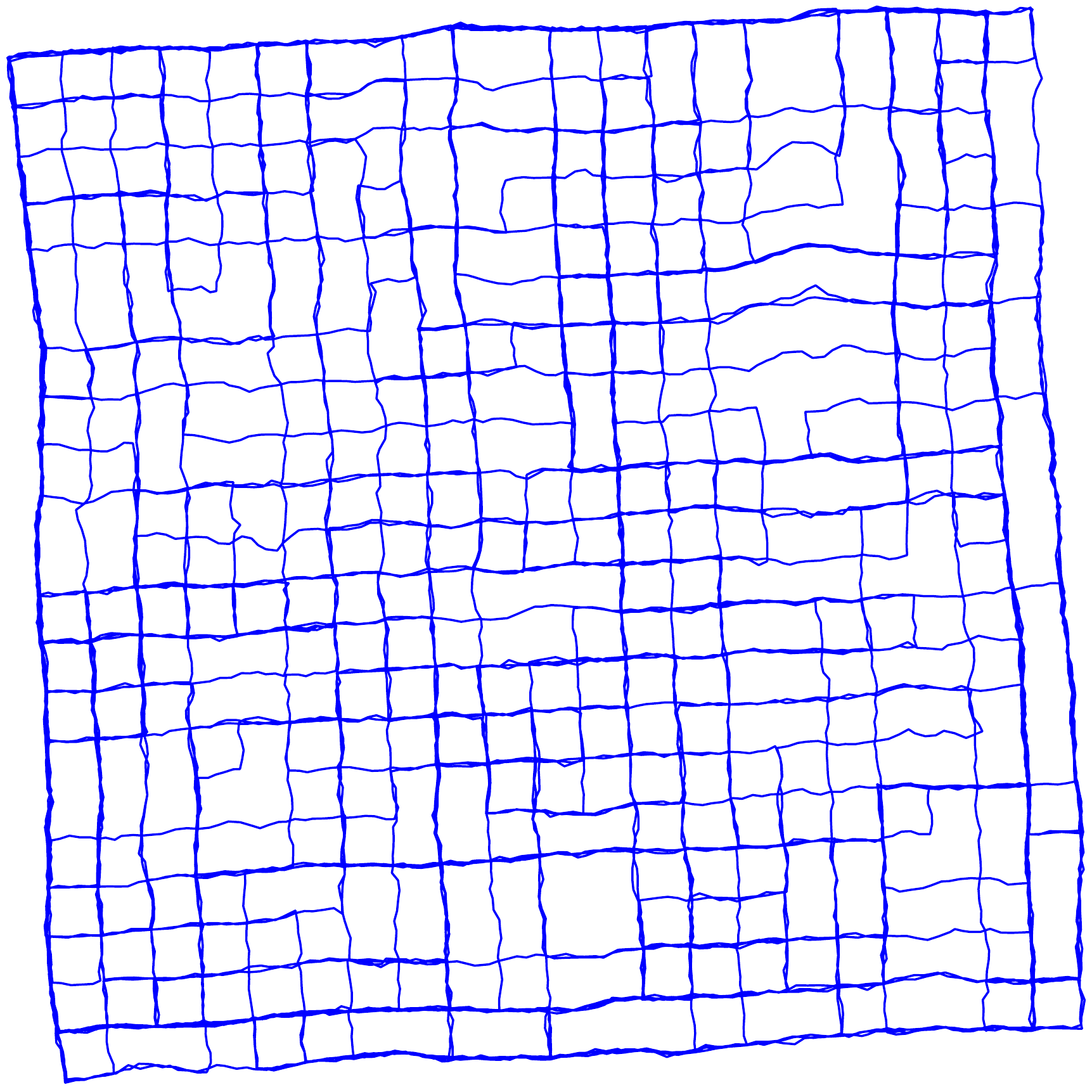} \\
			\etrans{4.71}{-1} & \etrans{2.46}{-1} \\
			\erot{1.51}{1} & \erot{1.22}{1} \\ 
		\end{tabular}
		\caption{Sample visualization of results with chordal relaxation under large odometry noise.}
		\label{fig:chordHard}
	\end{figure}
	\begin{figure}[t]	
		\centering
		\begin{subfigure}{0.4\textwidth}	
			\includegraphics[width = \linewidth] {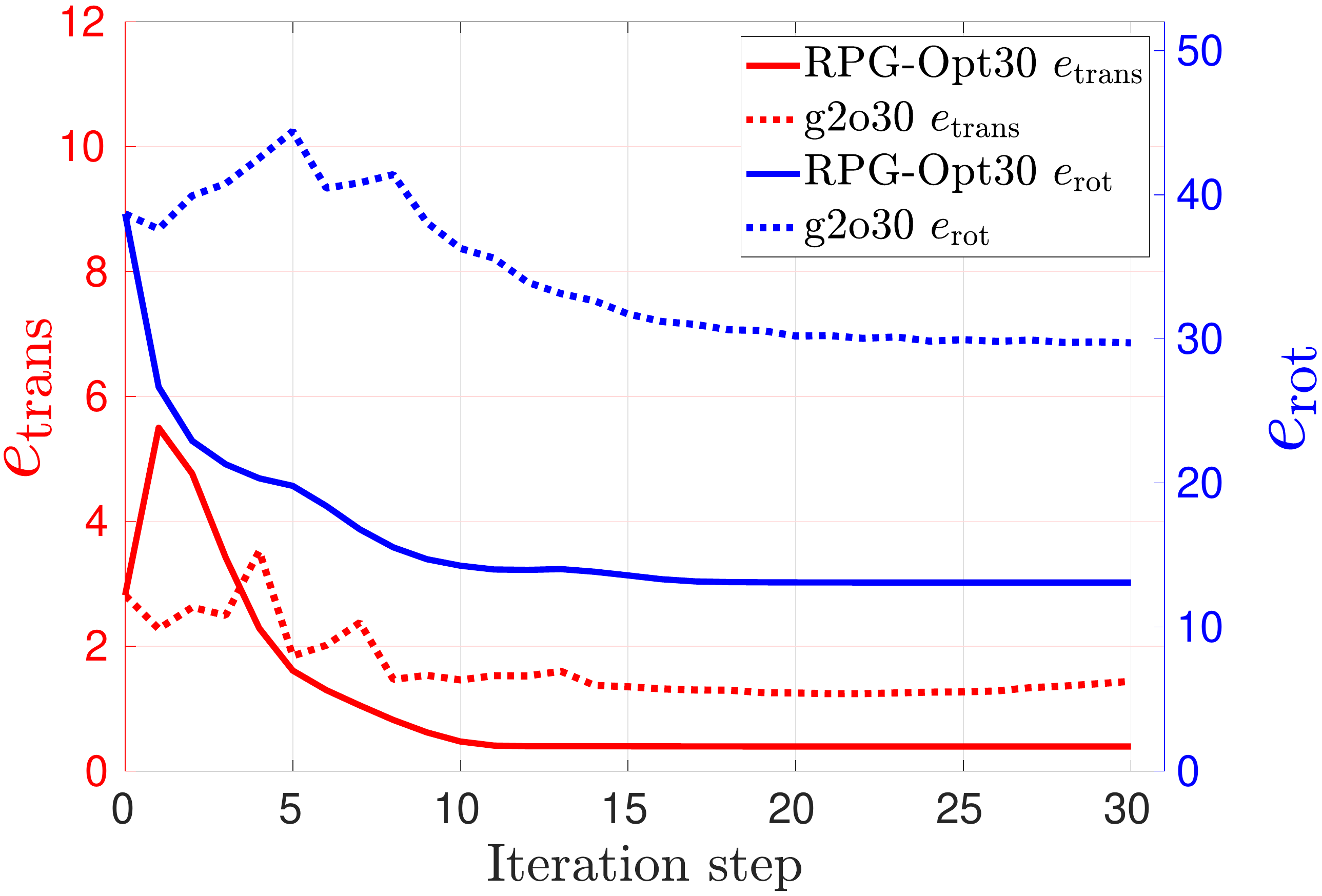}
			\caption{Change of RPE for each iteration.}
		\end{subfigure}	
		\begin{subfigure}{0.41\textwidth}	
			\includegraphics[width = 0.9\linewidth] {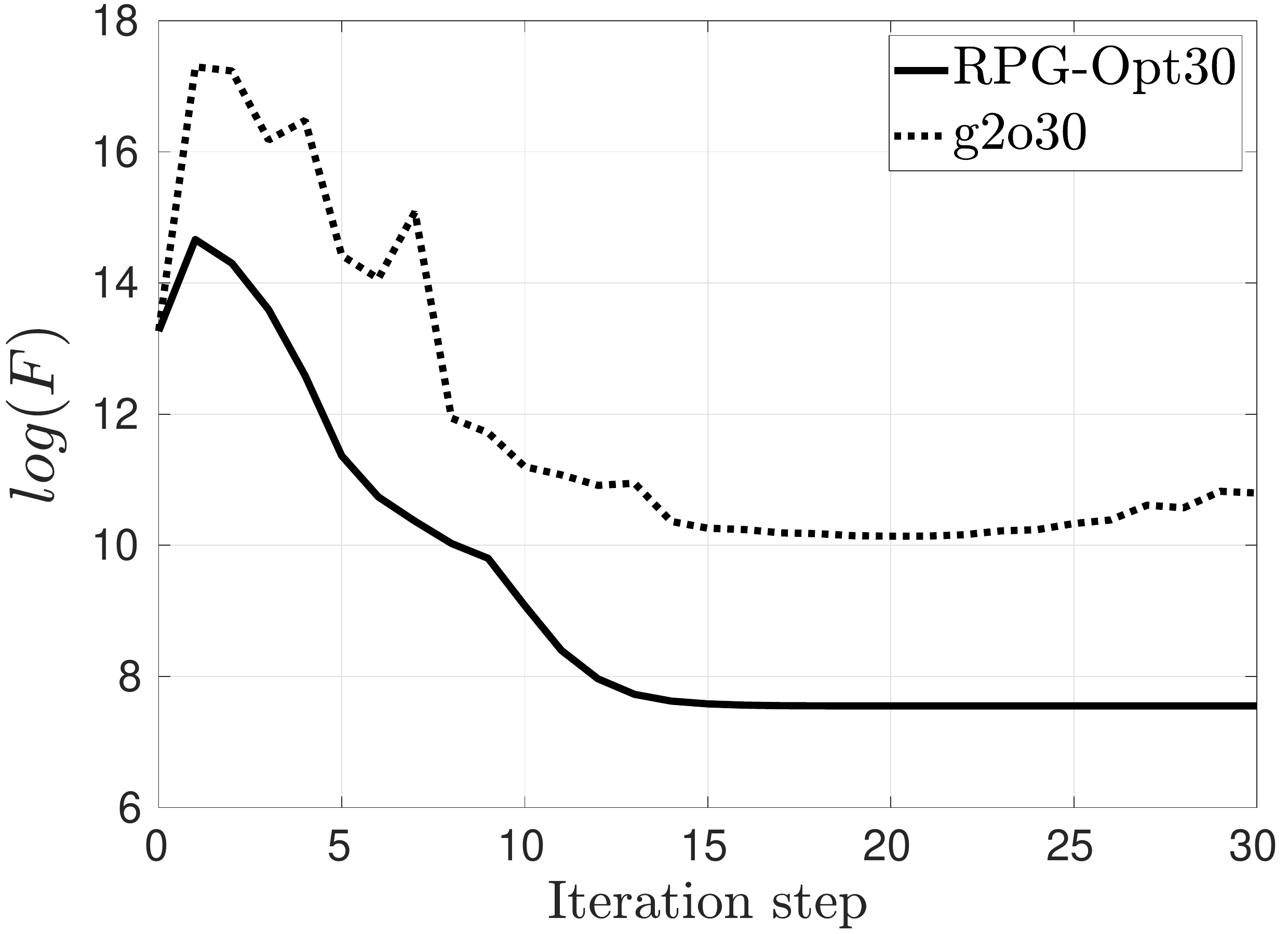}
			\caption{Change of cost function values defined in (\ref{eq:obj}).}
		\end{subfigure}	
		\caption{Convergence of RPG-Opt and g2o for \textit{M3500c+}.}
		\label{fig:convergence}
	\end{figure}
	
	\section{Evaluation} \label{sec:eva}
	We evaluate the proposed RPG-Opt based on publicly available planar pose graph data sets given in~\cite{carlone2014angular} and~\cite{carlone2014fast}, and compare it with three state-of-the-art pose graph optimization frameworks: g2o in~\cite{grisetti2010tutorial}, GTSAM in~\cite{dellaert2012factor} and iSAM in~\cite{Kaess08tro}. We choose to use Levenberg--Marquardt algorithm for iSAM and the Gauss--Newton method for GTSAM. In g2o, the optimization is performed using the Gauss--Newton algorithm. For most of the publicly available pose graph data sets, the uncertainty of the odometry  measurements is described by the information matrix for the state $[\,\theta,\,\vt_\Tx,\,\vt_\Ty\,]^\top$\,, with $\theta$ being the rotation angle and $[\,\vt_\Tx,\,\vt_\Ty\,]^\top$ the translation. In order to incorporate the estimated uncertainty, the logarithm map in (\ref{eq:logmap}) can be reformulated into the following form
	\begin{equation}
	\label{eq:state2t}
	\begin{aligned}
	&\quad\Log_\mathds{1}(\vz_{ij}^{-1}\boxplus\vx_i^{-1}\boxplus\vx_j)=\frac{1}{2}\bbmat 1&~~~0&0\\
	0 &~~~\beta &\alpha\\
	0 &-\alpha &\beta\ebmat\bbmat\delta\theta\\\delta\vt_\Tx\\ \delta \vt_\Ty\ebmat\,,\\
	\end{aligned}
	\end{equation}
	with $\alpha=\delta\theta/2$ and $\beta=\cos(\delta\theta/2)/\sinc(\delta\theta/2)$\,.
	Here, $\delta\theta$ and $[\delta\vt_\Tx\,,\delta \vt_\Ty]^\top$ is the rotation angle and translation term of the planar dual quaternion  $\vz_{ij}^{-1}\boxplus\vx_i^{-1}\boxplus\vx_j$\,, respectively. The equation in (\ref{eq:state2t}) is essentially a nonlinear transformation of the state $[\delta\theta,\delta\vt_\Tx,\delta\vt_\Ty]^\top$ provided by the data set. For each iteration step, we can assume that the odometry error is small, meaning  $\alpha\rightarrow0$ and $\beta\rightarrow1$\,. Therefore, the information matrix from the raw data set can be directly deployed as the on-tangent-plane information matrix $\bm{\Omega}_{ij}$ in the metric of (\ref{eq:state2t})\,. When the information matrices for odometry are not available, a typical way to formulate the cost function is to set $\bm{\Omega}_{ij}=\fI\in\R^{3\times3}$\,.
	
	In the following evaluations, we first test the proposed approach based on data sets under ordinary odometry noise level using the identity and real information matrix, respectively. Second, we synthesize data sets with additional noise and perform the evaluation w.r.t. the ground truth.  
	\subsection{Evaluation Under Ordinary Odometry Noise Level}
	We first evaluate the proposed approach using data sets with ordinary level of odometry noise. As no ground truth is available, we use the cost function of g2o for comparison (\cite{kummerle2011g}), namely 
	\begin{equation*}
	\begin{aligned}
	\sum_{\langle i,j\rangle\in\mathds{C}}\bigg\Vert 
	\bbmat
	\Log\,(\fR_{ij}^\top \, \fR_{i}^\top \, \fR_{j}) \\
	\fR_{ij}^\top\li(\vt_{ij} - \fR_{i}^\top(\vt_j - \vt_i)\ri) 
	\ebmat
	\bigg\Vert^2_{\bm{\Omega}_{ij}}\,,
	\end{aligned}
	\end{equation*}
	with $\fR\in\SOT$ being the ordinary rotation matrices and $\vt$ the translations. 
	
	Here, the logarithm map of rotation matrices can be derived from Lie algebra as given in~\cite{carlone2015initialization}. Table~\ref{tab:easy} shows the results for both identity and real information matrices (denoted as $\fI$ and $\bm{\Omega}$, respectively). Here, g2o and RPG-Opt iteration steps are fixed to be $10$\,, whereas GTSAM and iSAM use their default stopping criterion. Our approach, the RPG-Opt, reaches comparable accuracy as the state of the art.
	
	\subsection{Evaluation Based on Synthetic Data sets}
	We further synthesize two groups of data sets with known information matrices for the evaluation under large uncertainty of odometry measurements. For that, we add additional odometry noise to the data sets \textit{M3500s} and \textit{City10000} in~\cite{carlone2014angular}. Fig.~\ref{fig:hard} shows the results based on the relative pose error (RPE) for both translations and rotations, which are denoted as $e_\text{t}$ and $e_\text{r}$, respectively. The optimizations are initialized directly from the odometry measurements. Results from GTSAM using the Gauss--Newton approach are not listed because of nonconvergence. For all the sequences, the g2o and iSAM framework are prone to the local minima. However, the proposed approach shows the best accuracy and robustness against local minima, though no special initialization is performed. 
	
	For pose graph optimization under large uncertainty of odometry measurements, it is typical to equip the optimizer with an additional initialization block for better convergence. Therefore, we incorporate an initialization method based on chordal relaxation in~\cite{carlone2015initialization}. Fig.~\ref{fig:chordHard} visualizes evaluation results with chordal relaxation-based initialization using another two synthetic data sets based on the \textit{M3500} and \textit{City10000}. In this case, the g2o shows significant improvement, achieving the same converged accuracy as the proposed approach. As a summary, Table~\ref{tab:sum} shows the optimization results with chordal initialization for all the synthetic data sets w.r.t. the ground truth. For data sets \textit{M3500a+}, \textit{M3500b+}, \textit{M3500c+} and \textit{City10000a}, the proposed approach initialized with odometry measurements achieves the same errors as the other approaches using the chordal initialization. Results from iSAM are not available, because external initializations are not supported by the framework. We additionally list the results with direct odometry initialization for the RPG-Opt to show the convergence robustness under large uncertainties of the proposed method . 
	
	\subsection{Convergence}
	We further compare the convergence behavior of the proposed approach with g2o as shown in Fig.~\ref{fig:convergence}. Here, the sequence \textit{M3500c+} is used and the RPE for both translation and rotations are plotted in Fig.~\ref{fig:convergence}-(a). Fig.~\ref{fig:convergence}-(b) shows the convergence of the proposed Riemannian metric in (\ref{eq:obj}). The optimizations are initialized with odometry and the real information matrices are used. In both plots, the proposed approach shows faster convergence and better robustness against local minimum compared to g2o. This mainly results from the geometry-aware cost function formulation and the Riemannian Gauss--Newton of the proposed approach. 
	
	\section{Conclusion} \label{sec:conclusion}
	In this work, we have proposed a Riemannian approach for planar pose graph optimization problems on the manifold of dual quaternions. Here, the cost function is built upon the Riemannian metric of the planar dual quaternion manifold, based on which a Riemannian Gauss--Newton method is applied using the exponential retraction. Both the on-manifold MLE formulation and optimization are geometry-aware, which inherently consider the underlying nonlinear structure of the $\SET$ group. We further performed evaluations based on real-world and synthetic data sets with extra noise. 
	Compared with the state-of-the-art frameworks, the proposed approach gives equally accurate results under ordinary odometry noise and shows better accuracy and robustness under large uncertainty of odometry measurements.
	
	Based on the presented work, there is still much potential to exploit. The proposed Riemannian approach can be extended to the general dual quaternion-based spatial pose graph optimization. Also, numerous optimizers from the family of Riemannian optimization can be further applied for better convergence robustness and accuracy.

	\bibliographystyle{./ifacconf.bst}
	\bibliography{./ISASPublikationen,./bibliography_local}
\end{document}